\documentclass[10pt,twocolumn,letterpaper]{article}

\usepackage{cvpr}              

\usepackage{overpic}
\usepackage{enumitem} 
\usepackage{overpic} 
\usepackage{color}
\usepackage[dvipsnames]{xcolor}
\usepackage{wrapfig}

\usepackage{tikz}
\usepackage{comment}
\usepackage[font=small]{caption}
\usepackage{colortbl}
\usepackage{subcaption}
\usepackage{bbm}
\usepackage{color}

\makeatletter
\DeclareRobustCommand\onedot{\futurelet\@let@token\@onedot}
\def\@onedot{\ifx\@let@token.\else.\null\fi\xspace}

\def\eg{\emph{e.g}\onedot} 
\def\ie{\emph{i.e}\onedot}

\def\etal{\emph{et al}\onedot}
\makeatother

\definecolor{turquoise}{cmyk}{0.65,0,0.1,0.3}
\definecolor{purple}{rgb}{0.65,0,0.65}
\definecolor{dark_green}{rgb}{0, 0.5, 0}
\definecolor{orange}{rgb}{0.8, 0.6, 0.2}
\definecolor{red}{rgb}{0.8, 0.2, 0.2}
\definecolor{darkred}{rgb}{0.6, 0.1, 0.05}
\definecolor{blueish}{rgb}{0.0, 0.3, .6}
\definecolor{light_gray}{rgb}{0.7, 0.7, .7}
\definecolor{pink}{rgb}{1, 0, 1}
\definecolor{greyblue}{rgb}{0.25, 0.25, 1}

\DeclareMathOperator*{\argmin}{arg\,min}

\usepackage{blindtext}

\renewcommand{\paragraph}[1]{\vspace{1em}\noindent\textbf{#1}.}

\definecolor{JonYellow}{rgb}{1,1, 0.6}
\definecolor{JonOrange}{rgb}{1, 0.8, 0.6}
\definecolor{JonRed}{rgb}{1, 0.6, 0.6}

\usepackage{enumitem}
\setlist[itemize]{align=parleft,left=0pt..1em}

\usepackage{amsmath,amsfonts,bm}
\usepackage{mathtools}

\def\eqref#1{equation~\ref{#1}}

\def\1{\bm{1}}

\DeclareMathAlphabet{\mathsfit}{\encodingdefault}{\sfdefault}{m}{sl}
\SetMathAlphabet{\mathsfit}{bold}{\encodingdefault}{\sfdefault}{bx}{n}

\newcommand{\E}{\mathbb{E}}

\newcommand{\defeq}{\coloneqq}

\newcommand{\Eb}[2]{\E_{#1}\!\left[#2\right]}

\newcommand{\bI}{\mathbf{I}}

\newcommand{\bzero}{\mathbf{0}}

\newcommand{\bc}{\mathbf{c}}

\newcommand{\bx}{\mathbf{x}}

\newcommand{\bz}{\mathbf{z}}

\newcommand{\bepsilon}{{\boldsymbol{\epsilon}}}

\newcommand{\bP}{\mathbf{P}}

\definecolor{cvprblue}{rgb}{0.21,0.49,0.74}
\usepackage[pagebackref,breaklinks,colorlinks,citecolor=cvprblue]{hyperref}

\def\paperID{1893} 
\def\confName{CVPR}
\def\confYear{2024}

\title{Infinite Texture: Text-guided High Resolution Diffusion Texture Synthesis\vspace{-0.5em}}

\author{Yifan Wang$^{1}$\qquad Aleksander Hołyński$^{2,3}$\qquad Brian L. Curless$^{1}$\qquad Steven M. Seitz$^{1}$\\
$^1$University of Washington \hspace{1cm} $^2$UC Berkeley \hspace{1cm} $^3$Google Research
}

\newcommand{\model}{Infinite Texture\xspace}

\begin{document}
\twocolumn[{%
    \renewcommand\twocolumn[1][]{#1}%
    \maketitle
    \begin{center}    
    \centering
    \captionsetup{type=figure}
    \includegraphics[width=\linewidth]{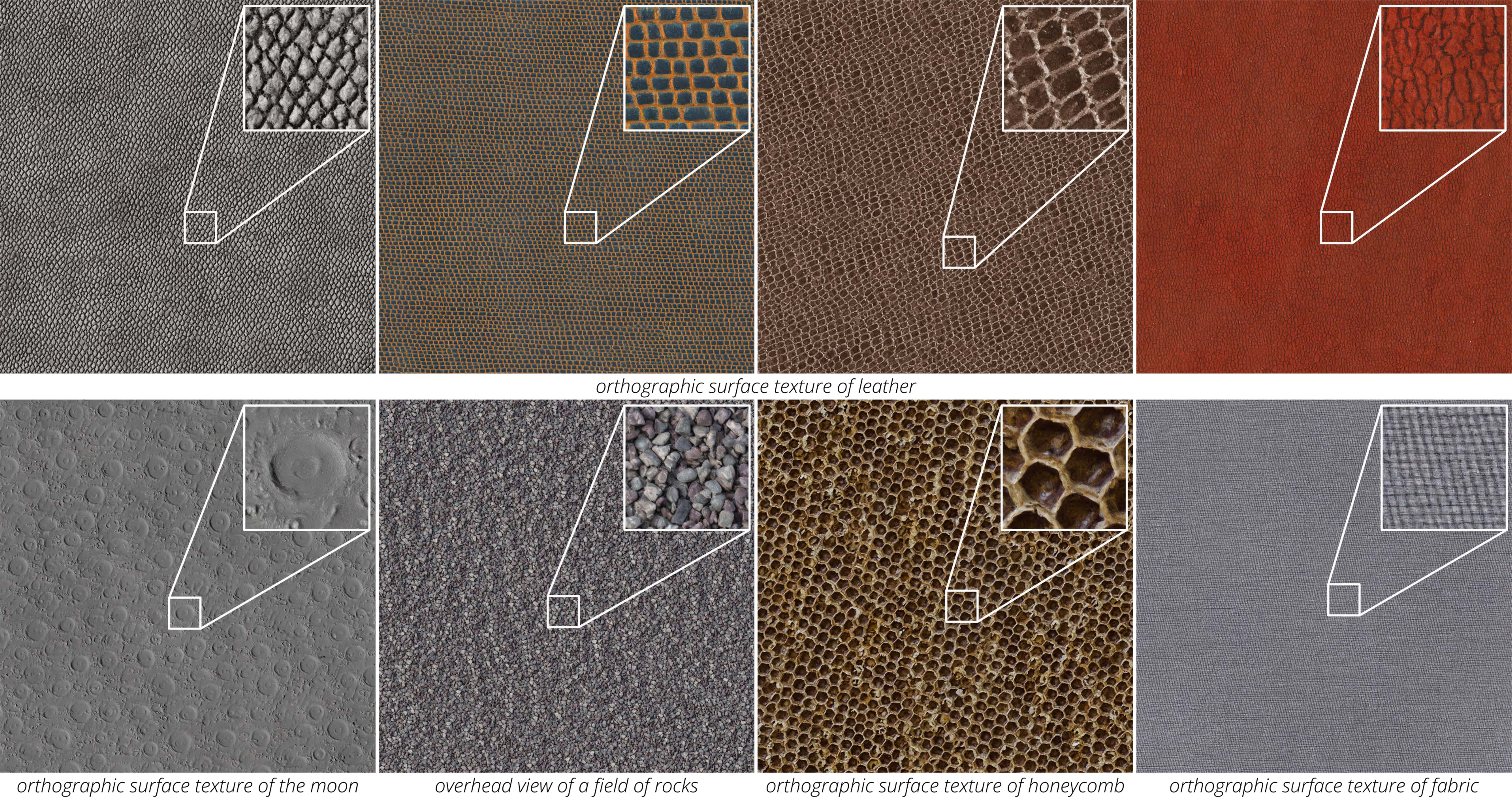}
    \captionof{figure}{
    \model generates arbitrarily large (examples here are 85MP), high-quality textures from text prompts. 
    Given a single text prompt as input, \model generates a diverse collection of textures (leather on the top row).
    Our method succeeds in reproducing both statistically periodic textures (fabric on the bottom row) and challenging ones with depth variations (honeycomb on the bottom row). 
    Close-up views of the samples are depicted within the white boxes. 
    }
    \label{fig:teaser}
\end{center}%
}]

\begin{abstract}
    We present \model, a method for generating arbitrarily large texture images from a text prompt.  Our approach fine-tunes a diffusion model on a single texture, and learns to embed that statistical distribution in the output domain of the model.
    We seed this fine-tuning process with a sample texture patch, which can be optionally generated from a text-to-image model like DALL-E~2. 
    At generation time, our fine-tuned diffusion model is used through a score aggregation strategy to generate output texture images of arbitrary resolution on a single GPU. We compare synthesized textures from our method to existing work in patch-based and deep learning texture synthesis methods. We also showcase two applications of our generated textures in 3D rendering and texture transfer.
\end{abstract}

\section{Introduction}
Textures, defined as statistically repeating image content, are fundamental primitives in computer graphics. They effectively describe a wide variety of surfaces, ranging from tree bark to human skin, capturing fine surface details. This capability enables the efficient generation of photorealistic imagery, surpassing what can be achieved with geometry alone. Realistic textures can significantly enhance the overall visual experience.

Texture assets are often created by artists, either using manual design tools or finding and capturing reference content.
Developments in texture synthesis~\cite{efros1999texture,turk2001texture,wei2000fast,efros2001image} propose to simplify this process by generating larger, randomly varying samples of a texture, from a smaller reference patch. 
Texture synthesis methods have evolved significantly over the years, from statistical or patch-based models~\cite{efros1999texture,turk2001texture,wei2000fast,efros2001image} to more modern deep-learning solutions~\cite{gatys2015texture,bergmann2017learning,zhou2018non}.
Most existing methods have been demonstrated only on low resolution textures, \ie, up to $512 \times 512$. 
Furthermore, all these methods involve an arduous process, as an artist must still capture or design an input reference texture.   

In this paper, we focus on an alternative guiding signal for texture synthesis: text. Recently, text-based image generation has seen remarkable progress with the introduction of diffusion models~\cite{dhariwal2021diffusion, rombach2021highresolution,ramesh2022hierarchical}, allowing generative models to synthesize high-quality images based on arbitrary text prompts. Still, producing high-quality textures through this interface is non-trivial: (1) choosing the correct prompt to get a texture-like image is often challenging, (2) most models cannot generate images beyond $1024 \times 1024$, and textures are often needed at higher resolutions, and (3) a text interface does not easily enable generation of stochastic permutations of a particular texture.

To address these challenges, we introduce \model, a prompt-based approach for high-resolution texture generation using a text-to-image diffusion model. As demonstrated in Fig.~\ref{fig:teaser}, our method takes a text prompt as input, and generates a varying collection of high-resolution, high-quality textures.
Our method uses a text-to-image diffusion model~\cite{ramesh2022hierarchical} to synthesize a reference texture, which is then used to fine-tune a diffusion model for texture synthesis. We select a set of prefixes for input prompts to ensure DALL-E~2~\cite{ramesh2022hierarchical} produces texture-like images.
The diffusion model is trained on patches of the reference texture to learn its statistical properties and is able to generate novel samples with stochastic permutations of the reference texture. 
Since diffusion models are computationally expensive, we propose a score aggregation strategy that allows a diffusion model to progressively synthesize an arbitrarily large texture. 

We demonstrate the utility of \model for creating high-quality textures from text prompts through a diverse collection of synthesized results, and show that 
our method performs favorably when compared with both traditional and data-driven texture synthesis methods. Additionally, we showcase two applications enabled by our generated textures: 3D rendering, and re-texturing real images. 

In summary, our main contributions are:
\begin{itemize}
    \item a practical pipeline for generating high-quality and high-resolution textures purely from text descriptions, involving:
    \begin{itemize}
    \item a novel approach for generating a variety of high-resolution, high-quality textures from a reference texture,
    \item a strategy for generating arbitrarily large textures from a pretrained diffusion model (\ie, output texture sizes that significantly exceed those of any existing method).
    \end{itemize}
    \item a diffusion-based approach for re-texturing an input image with the textures generated by \model.
    
\end{itemize}

\section{Related Work}

\paragraph{Texture synthesis}
Texture synthesis has been an active research area in computer graphics for several decades. Traditional approaches for texture synthesis can be broadly categorized into two groups: pixel-based synthesis and patch-based synthesis. Efros and Leung~\cite{efros1999texture} proposed to synthesize textures by gradually growing the map from the initial texture, assigning the output pixels one by one in an inside-out, onion-layer fashion. 
Despite its elegance and simplicity, this method is relative slow and subject to non-uniform pattern distribution. Wei~\etal~\cite{wei2000fast} proposed a simple pixel-based texture synthesis algorithm based on fixed neighborhood search. The quality and speed of pixel-based approaches can be improved by synthesizing patches rather than pixels. Praun~\etal~\cite{praun2000lapped} repeatedly pastes new patches over existing regions. By using patches with irregular shapes, this method takes advantage of the human visual system and its perception of texutre mapping effects, making texture synthesis work 
 surprisingly well for stochastic textures. Liang~\etal~\cite{liang2001real} take a different approach by using a blending algorithm for overlapping regions. Efros and Freeman~\cite{efros2001image} use dynamic programming to find an optimal path to cut through the overlapped regions, and this idea is improved by~\cite{kwatra2003graphcut} via graph cut. Kaspar~\etal~\cite{kaspar2015self} further extends Texture Optimization~\cite{kwatra2003graphcut} by making its various parameters and weights to be self-tunable.

Another line of works use Cellular Automata (CA) for texture generation~\cite{turk1991generating,witkin1991reaction}. Recent work by Niklasson~\etal~\cite{niklasson2021self} proposed a learning-based Texture Neural Cellular Automata model where the CA update rule is parameterised with a small neural network. This idea is further investigated in~\cite{mordvintsev2021mu} by making the model more compact.

Texture synthesis has seen significant progress with the advent of deep learning-based methods. Gatys~\etal~\cite{gatys2015texture} synthesize new textures by minimizing the Gram loss with the original texture. Followup works~\cite{ulyanov2016texture,ulyanov2017improved,li2017diversified} speed up the synthesis by adopting a feed-forward generative network. Jetchev~\etal~\cite{jetchev2016texture} utilize GANs to generate texture patches from random noise of the same size. Bergmann~\etal~\cite{bergmann2017learning} extends this idea by by introducing a periodic function into the input noise, which enables synthesizing high-quality periodic textures. Zhou~\etal~\cite{zhou2018non} train a generative network to double the spatial extent of texture blocks for synthesizing non-stationary textures. A GAN-based texture synthesis method was used by Verbin and Zickler~\cite{Verbin_2020_CVPR} for estimating surface shape using texture cues, and Verbin~\etal~\cite{verbin2} presented a mathematical formulation for the uniqueness of the solution to the surface recovery problem.

Traditional texture synthesis methods are largely slow and inefficient. Deep learning-based methods are restricted by the receptive field and fail to learn the extreme low- or high-frequncy signal in the texture. We introduce a novel method to leverage image priors from diffusion models and synthesize arbitrarily large, high-quality textures on a single GPU.

\paragraph{Text-to-image synthesis}
Generating realistic images from textual descriptions is a challenging task. Early attempts~\cite{mansimov2015generating,reed2016generative,tao2022df,xu2018attngan,zhang2017stackgan,zhu2019dm} we limited as they employed text-conditional GANs on specific domains~\cite{welinder2010caltech} and their datasets held closed-world assumptions ~\cite{lin2014microsoft}. However, recent advances in diffusion models~\cite{dhariwal2021diffusion,ho2020denoising} and large-scale language encoders~\cite{radford2021learning,raffel2020exploring} have greatly improved text-to-image synthesis, enabling these models to be conditioned on an open-world vocabulary of arbitrary text descriptions. Prominent diffusion models, such as GLIDE~\cite{nichol2021glide}, DALL-E 2~\cite{ramesh2022hierarchical}, and Imagen~\cite{saharia2022photorealistic}, produce photorealistic outputs with the aid of a pretrained language encoder~\cite{radford2021learning,raffel2020exploring}. Although diffusion models have demonstrated unprecedented image synthesis ability, the iterative image sampling processes in the denoising step is often time-consuming. To accelerate the sampling, several methods~\cite{lu2022dpm,meng2022distillation,salimans2022progressive,song2020denoising} propose solutions for reducing the number of sampling steps. Latent Diffusion Models~\cite{rombach2021highresolution} also greatly speeds up the sampling by sampling in a low-dimensional latent space instead of pixel space. 

Since off-the-shelf diffusion models are trained to produce arbitrary images, they are often not particularly well suited for synthesizing stochastically varying periodic textures. We propose a new way to help diffusion models produce textural content by fine-tuning a model to learn texture statistics. Diffusion models are also difficult to use for synthesizing high resolution images due to the memory constraint. For this, we further introduce a strategy for generating spatially-consistent and large textures at inference time.

\begin{figure*}[!t]
    \begin{center}

    \begin{subfigure}[b]{0.32\linewidth}
    \includegraphics[width=\linewidth]{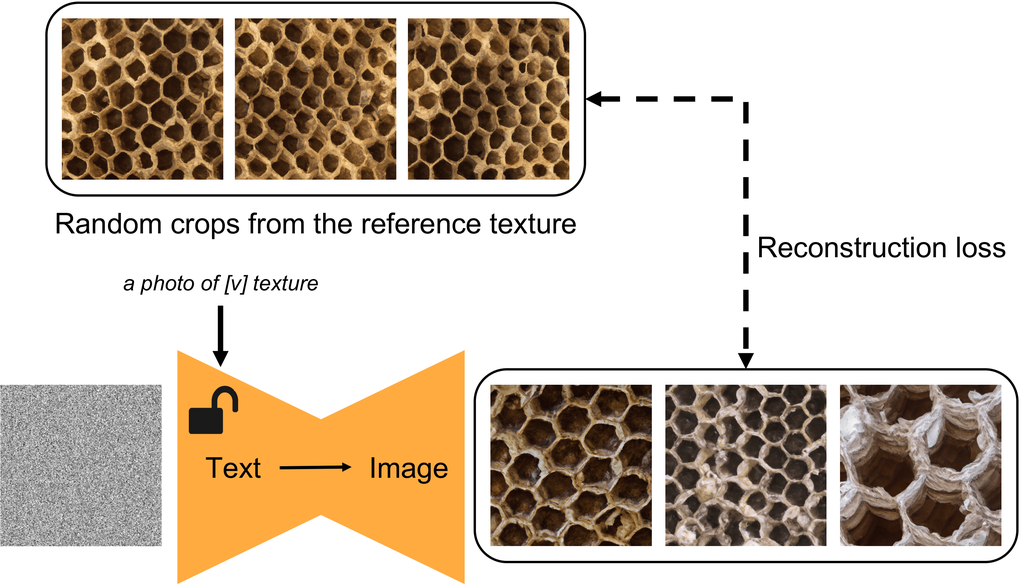} %
    \caption*{Per-texture Fine-tuning}
    \end{subfigure}
    \begin{subfigure}[t]{0.64\linewidth}
    \includegraphics[width=\linewidth]{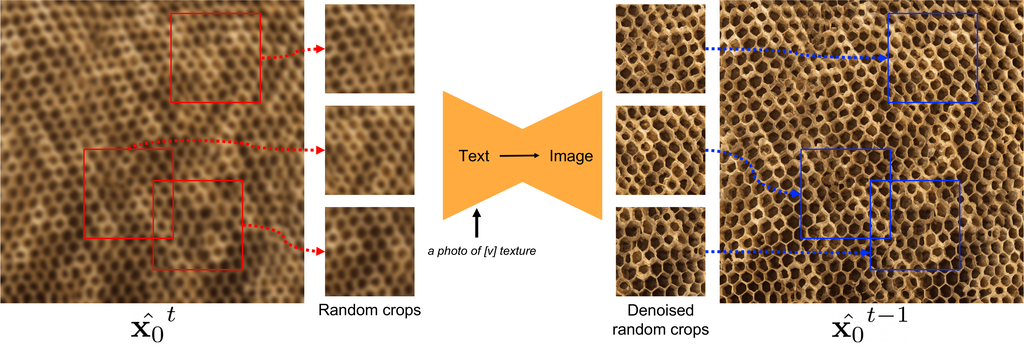} %
    \caption*{Texture Synthesis}
    \end{subfigure}
    
    \end{center}

    \caption{\textbf{Overview.} \model consists of three stages: (1) Generating a reference texture image from the text prompt; (2) Fine-tuning a diffusion model to learn the texture statistics of a particular reference texture image. We train the model using random crops of the reference texture image along with a unique identifier. The text encoder is also trained during this stage; and (3) We use the trained diffusion model to synthesize arbitrarily large textures: at every timestep of diffusion, we denoise small random patches and combine their estimates by taking the average noise estimate in overlapping regions. The resulting noise estimate is used to perform a DDIM sampling step.}
    \label{fig:overview}
\end{figure*}

\section{Method}

Our objective is to generate a diverse collection of high-resolution and high-quality texture samples based on an input text prompt. \model achieves this in three stages, as shown in Fig.~\ref{fig:overview}: (1) generating a reference texture image from the text prompt, (2) fine-tuning a diffusion model to learn the statistical distribution of the texture, and (3) combining the output of the diffusion model to synthesize a high-resolution texture. We next provide some background on diffusion models (Sec.~\ref{sec:diffusion}), our fine-tuning technique to learn the texture statistics (Sec.~\ref{sec:dreambooth}), and our novel approach for texture synthesis (Sec.~\ref{sec:multidiffison}).

\subsection{Diffusion Models}
\label{sec:diffusion}
Diffusion models are probabilistic generative models that are trained to learn a data distribution by gradually denoising a variable sampled from a Gaussian distribution. Specifically, we are interested in a pretrained text-to-image diffusion model $\hat\bx_\theta$ that, given an initial noise map $\bepsilon \sim \mathcal{N}(\bzero, \bI)$ and a conditioning vector $\bc=\Gamma(\bP)$, generates an image $\bx_{\text{gen}}=\hat\bx_\theta(\bepsilon, \bc)$. The conditioning vector is generated using a text encoder $\Gamma$ and a text prompt $\bP$. The pretrained model is trained to minimize a squared error loss to denoise a latent code $\bz_t \coloneqq \alpha_t \bx + \sigma_t \bepsilon$ as follows:

\begin{equation}
    \Eb{\bx,\bc,\bepsilon,t}{w_t \|\hat\bx_\theta(\alpha_t \bx + \sigma_t \bepsilon, \bc) - \bx \|^2_2}
    \label{eq:diffusion}
\end{equation}

\noindent where $\bx$ is the ground-truth image, $\bc$ is a conditioning vector (e.g., obtained from a text prompt), and $\alpha_t, \sigma_t, w_t$ are terms that control the noise scheduler and sample quality, and are functions of the diffusion process time $t \sim \mathcal{U}([0, 1])$. At inference time, the diffusion model is sampled by iteratively denoising $\bz_{t_1} \sim \mathcal{N}(\bzero, \bI)$ using either the deterministic DDIM~\cite{song2020denoising} or the stochastic ancestral sampler~\cite{ho2020denoising}. Intermediate points $\bz_{t_1}, \dotsc, \bz_{t_T}$, where $1 = t_1 > \cdots > t_T = 0$, are generated, with decreasing noise levels. These points, $\hat{\bx}^t_0 \defeq \hat\bx_\theta(\bz_t, \bc)$, are functions of the $\bx$-predictions.

Recent state-of-the-art text-to-image diffusion models use cascaded diffusion models in order to generate high-resolution images from text~\cite{saharia2022photorealistic,ramesh2022hierarchical}. Specifically, DALL-E 2~\cite{ramesh2022hierarchical} uses a base text-to-image model with $64 \times 64$ output resolution, and two unconditional super-resolution (SR) models $64\times 64 \rightarrow 256\times 256$ and $256\times 256 \rightarrow 1024\times 1024$. We use DALL-E~2 to generate a $1024\times 1024$ reference texture image from the text prompt.

\subsection{Per-Texture Fine-tuning}
\label{sec:dreambooth}
The aforementioned diffusion models~\cite{ramesh2022hierarchical,rombach2021highresolution} have shown unprecedented capabilities due to their ability to synthesize high-quality and diverse images based on text prompts. One of the main advantages of such model is strong image priors learned from a large collection of images that the model is initially trained on. While these models are capable of synthesizing interesting and high-quality images, they still lack the ability to reproduce the statistical distribution of a given reference texture image or to synthesize novel variations of the same texture. 

Wang \etal~\cite{wang2022pretraining} have demonstrated that a fine-tuned diffusion model outperforms a model trained from scratch for image translation tasks, especially when paired training data is limited. 
Following~\cite{wang2022pretraining,zhou2018non}, we choose to fine-tune a diffusion model for each reference texture image in order to learn the specific statistics associated with that texture.
We initialize our model with weights of a pretrained Stable Diffusion v2~\cite{rombach2021highresolution} checkpoint, leveraging its strong image priors. To ensure consistency in the output texture, we follow~\cite{ruiz2022dreambooth} and fine-tune the diffusion model with a unique identifier.

The pretrained Stable Diffusion is trained on a large and diverse dataset. However, we only require priors from a small portion of the training data to enable texture synthesis. We fine-tune the pretrained Stable Diffusion checkpoint to purposefully overfit on our reference texture image.
Given a reference texture image of $1024 \times 1024$, we take random crops of $768 \times 768$ and fine-tune the diffusion model on these texture patches. We overfit the model because we only want to inherit image priors, not semantic priors from the pretrained Stable Diffusion. For all patches, we use the text prompt $\bP$ of ``a photo of [identifier] texture'' to prevent any semantic prior. To further disentangle the semantic prior with image prior, the text encoder $\Gamma$ is also fine-tuned with the diffusion model $\hat\bx_\theta$. We use the same training loss as in Eq.~\ref{eq:diffusion} to fine-tune the diffusion model in an end-to-end manner. Weights for both the text encoder and the UNet are updated together at every iteration.

\begin{figure*}[!htbp]
    \begin{center}

    \begin{subfigure}[t]{0.07\linewidth}
    \includegraphics[width=\linewidth]{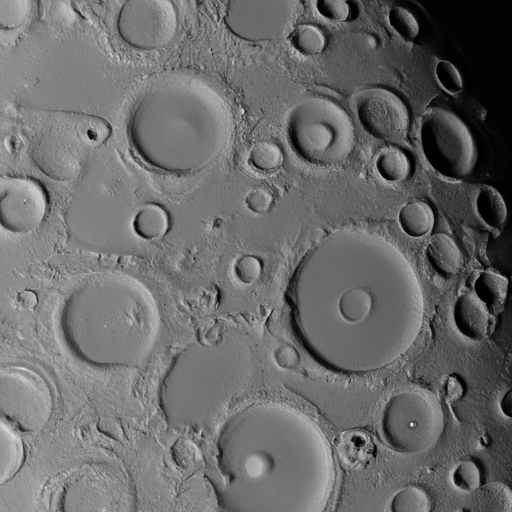} %
    \end{subfigure}
    \begin{subfigure}[t]{0.18\linewidth}
    \includegraphics[width=\linewidth]{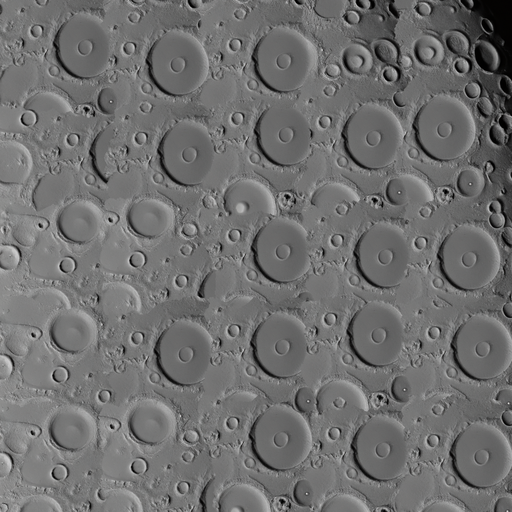} %
    \end{subfigure}
    \begin{subfigure}[t]{0.18\linewidth}
    \includegraphics[width=\linewidth]{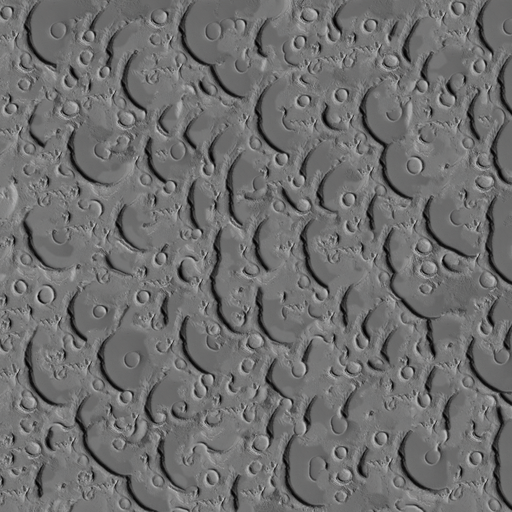} %
    \end{subfigure}
    \begin{subfigure}[t]{0.18\linewidth}
    \includegraphics[width=\linewidth]{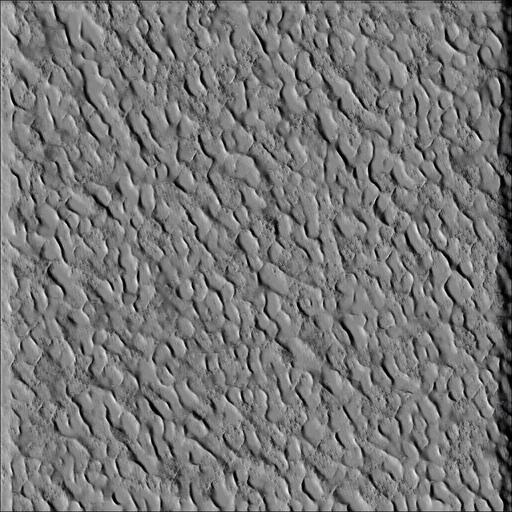} %
    \end{subfigure}
    \begin{subfigure}[t]{0.18\linewidth}
    \includegraphics[width=\linewidth]{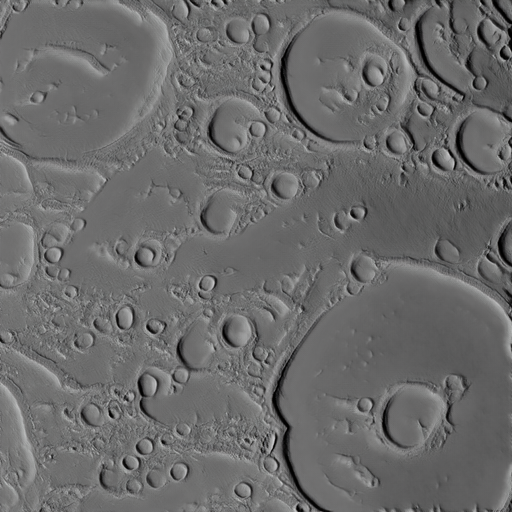} %
    \end{subfigure}
    \begin{subfigure}[t]{0.18\linewidth}
    \includegraphics[width=\linewidth]{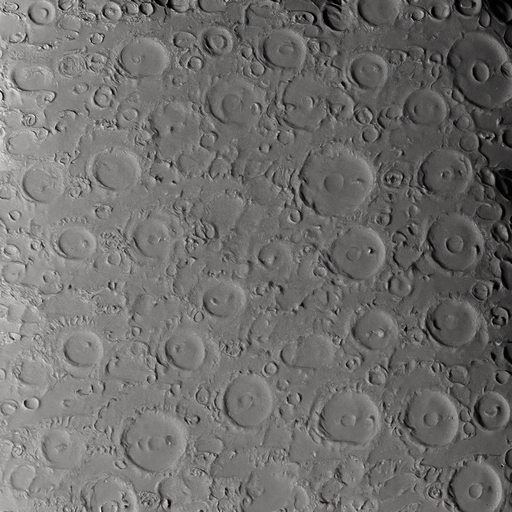} %
    \end{subfigure}\\

    \begin{subfigure}[t]{0.07\linewidth}
    \includegraphics[width=\linewidth]{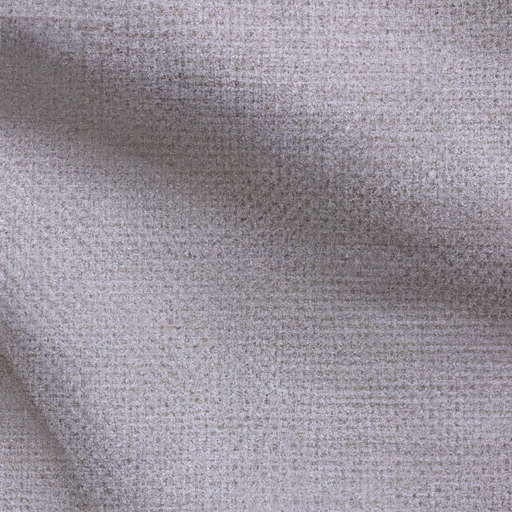} %
    \end{subfigure}
    \begin{subfigure}[t]{0.18\linewidth}
    \includegraphics[width=\linewidth]{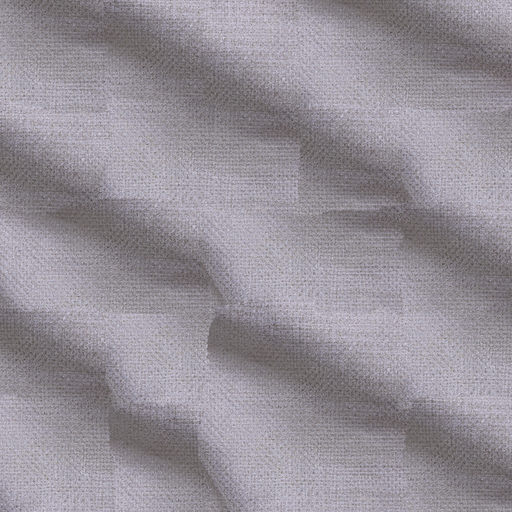} %
    \end{subfigure}
    \begin{subfigure}[t]{0.18\linewidth}
    \includegraphics[width=\linewidth]{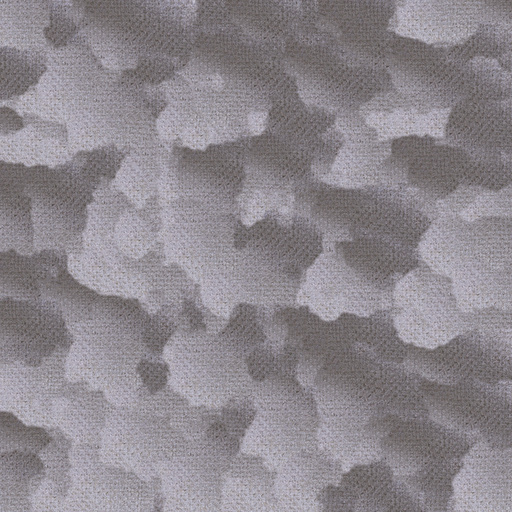} %
    \end{subfigure}
    \begin{subfigure}[t]{0.18\linewidth}
    \includegraphics[width=\linewidth]{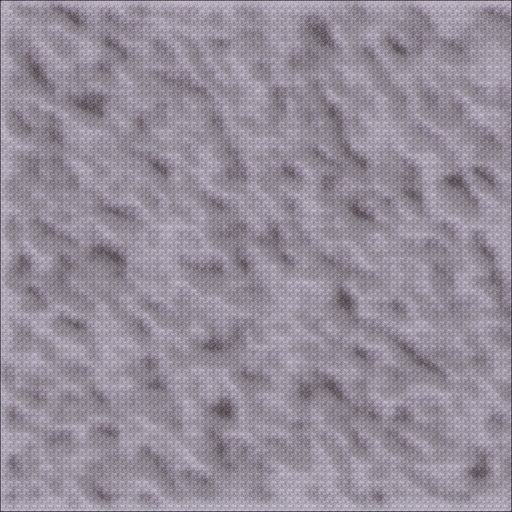} %
    \end{subfigure}
    \begin{subfigure}[t]{0.18\linewidth}
    \includegraphics[width=\linewidth]{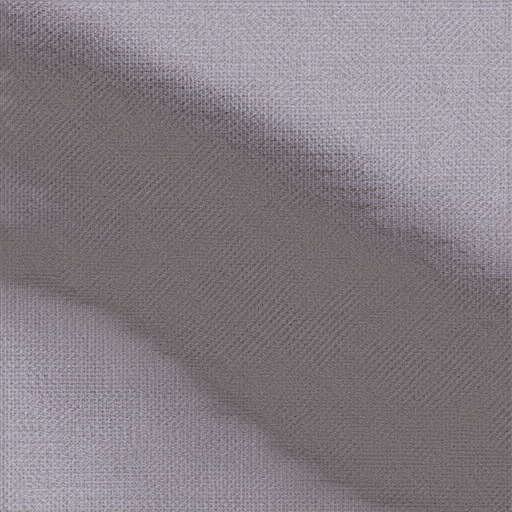} %
    \end{subfigure}
    \begin{subfigure}[t]{0.18\linewidth}
    \includegraphics[width=\linewidth]{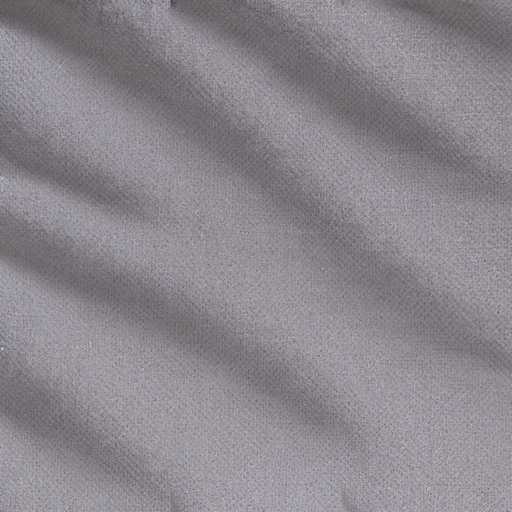} %
    \end{subfigure}\\

    \begin{subfigure}[t]{0.07\linewidth}
    \includegraphics[width=\linewidth]{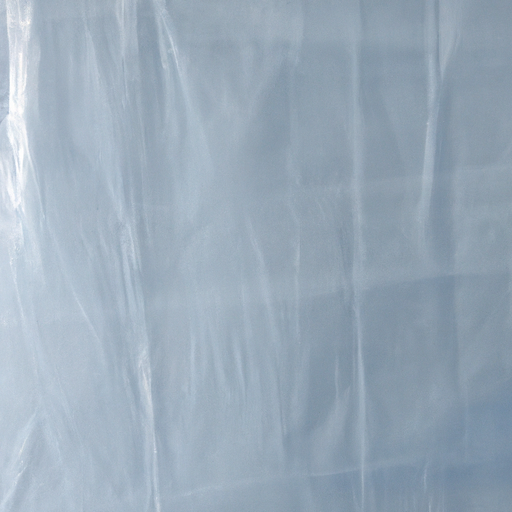} %
    \end{subfigure}
    \begin{subfigure}[t]{0.18\linewidth}
    \includegraphics[width=\linewidth]{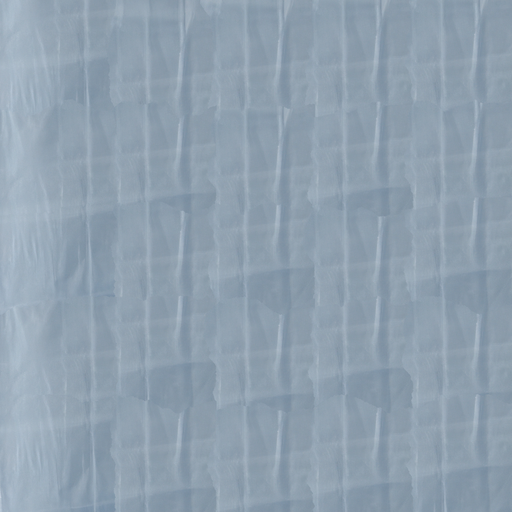} %
    \end{subfigure}
    \begin{subfigure}[t]{0.18\linewidth}
    \includegraphics[width=\linewidth]{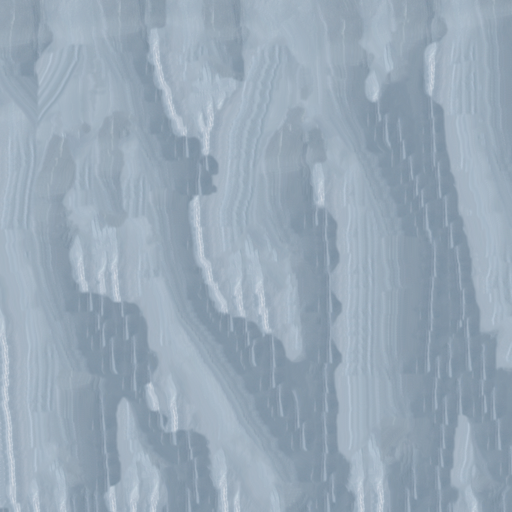} %
    \end{subfigure}
    \begin{subfigure}[t]{0.18\linewidth}
    \includegraphics[width=\linewidth]{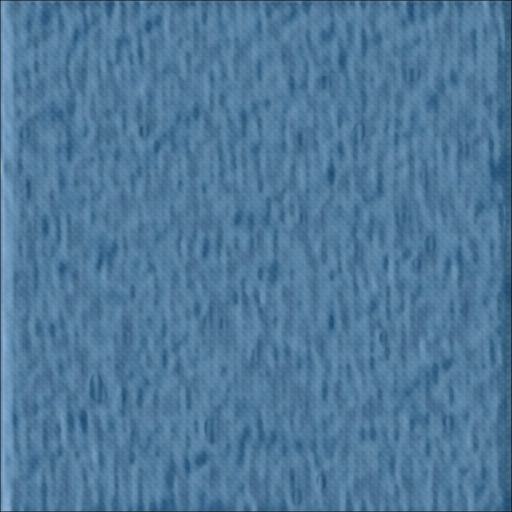} %
    \end{subfigure}
    \begin{subfigure}[t]{0.18\linewidth}
    \includegraphics[width=\linewidth]{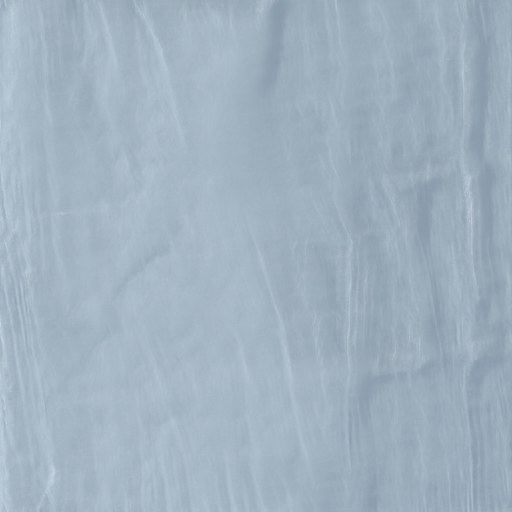} %
    \end{subfigure}
    \begin{subfigure}[t]{0.18\linewidth}
    \includegraphics[width=\linewidth]{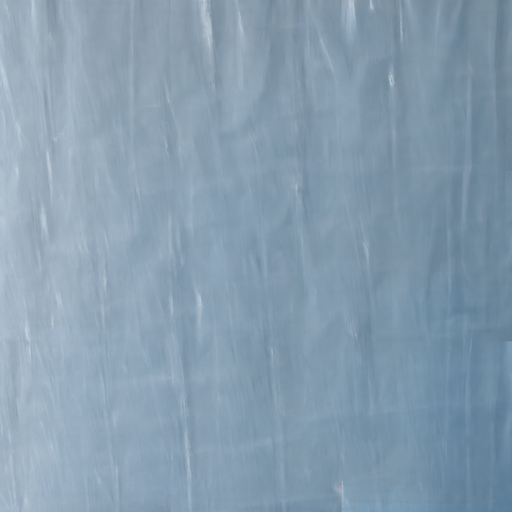} %
    \end{subfigure}\\

    \begin{subfigure}[t]{0.07\linewidth}
    \includegraphics[width=\linewidth]{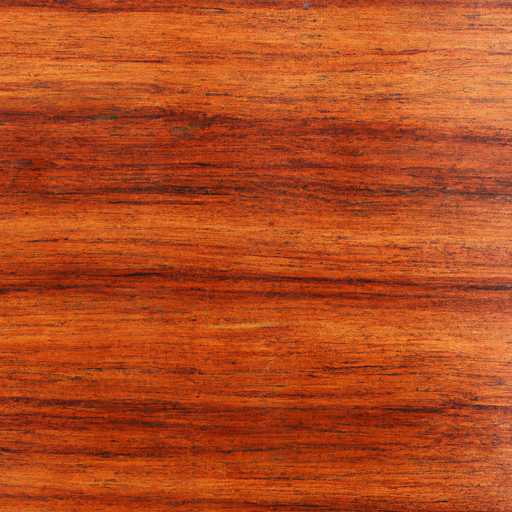} %
    \caption{Input}
    \end{subfigure}
    \begin{subfigure}[t]{0.18\linewidth}
    \includegraphics[width=\linewidth]{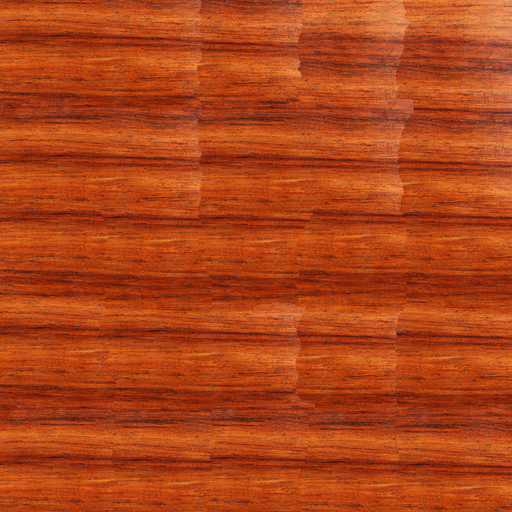} %
    \caption{Image Quilting~\cite{efros2001image}}
    \end{subfigure}
    \begin{subfigure}[t]{0.18\linewidth}
    \includegraphics[width=\linewidth]{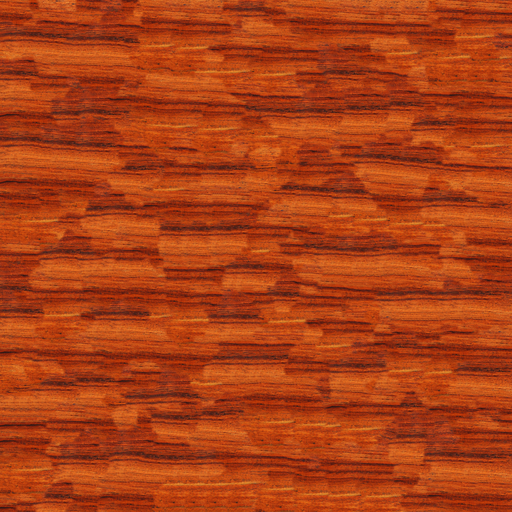} %
    \caption{STTO~\cite{kaspar2015self}}
    \end{subfigure}
    \begin{subfigure}[t]{0.18\linewidth}
    \includegraphics[width=\linewidth]{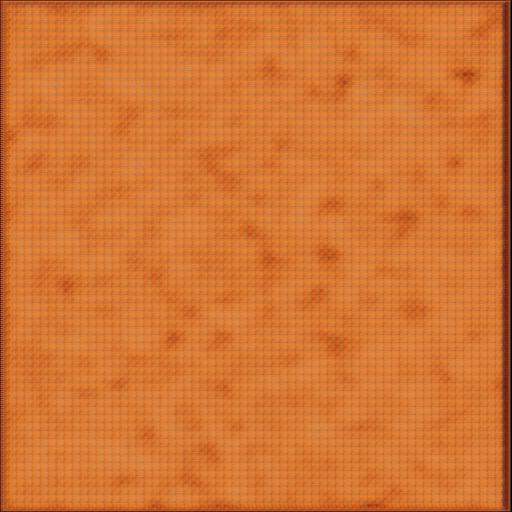} %
    \caption{PSGAN~\cite{bergmann2017learning}}
    \end{subfigure}
    \begin{subfigure}[t]{0.18\linewidth}
    \includegraphics[width=\linewidth]{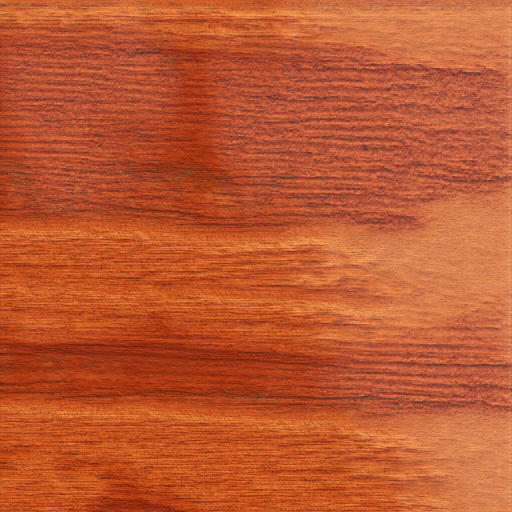} %
    \caption{NSTS~\cite{zhou2018non}}
    \end{subfigure}
    \begin{subfigure}[t]{0.18\linewidth}
    \includegraphics[width=\linewidth]{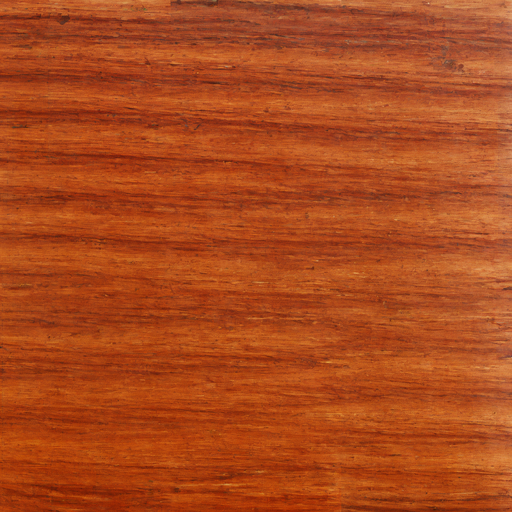} %
    \caption{Ours}
    \end{subfigure}

    \vspace{-5pt}

    \caption{\textbf{Comparisons with baseline methods in texture synthesis.} We use texture images generated by DALL-E~2~\cite{ramesh2022hierarchical} as input exemplar textures.
    \model stands out by generating the most consistent textures with variations. In contrast, image quilting~\cite{efros2001image} often results in repetitive patterns due to its tiling strategy. STTO~\cite{kaspar2015self} tends to converge to solutions where a smooth patch is repeated over and over. PSGAN~\cite{bergmann2017learning} struggles to capture the high-frequency signal effectively with its periodic signal generator. Non-stationary texture synthesis~\cite{zhou2018non} falls short in producing textures with variations due to only trained on small patches.}
    \label{fig:qualitative}

    \vspace{-20pt}

    \end{center}
\end{figure*}
\begin{figure}[t]
    \begin{center}

    \begin{subfigure}[t]{0.12\linewidth}
    \includegraphics[width=\linewidth]{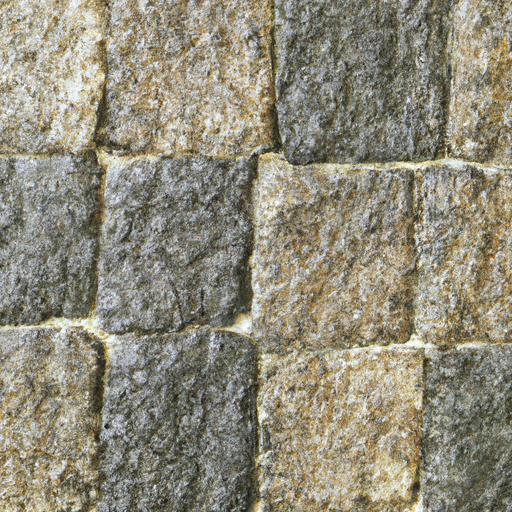} %
    \end{subfigure}
    \begin{subfigure}[t]{0.28\linewidth}
    \includegraphics[width=\linewidth]{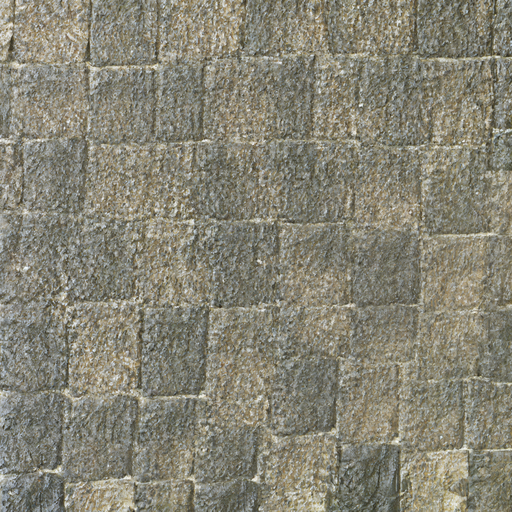} %
    \end{subfigure}
    \begin{subfigure}[t]{0.28\linewidth}
    \includegraphics[width=\linewidth]{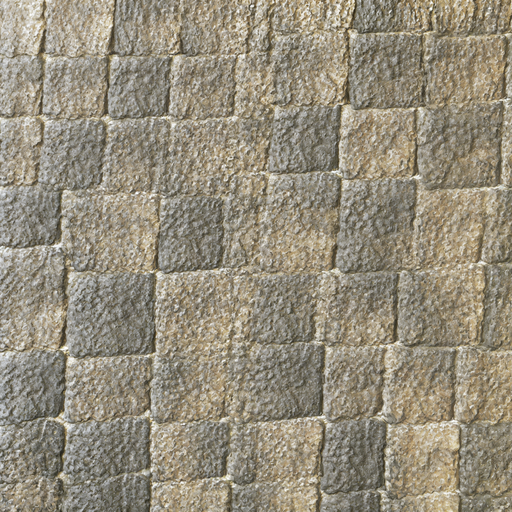} %
    \end{subfigure}
    \begin{subfigure}[t]{0.28\linewidth}
    \includegraphics[width=\linewidth]{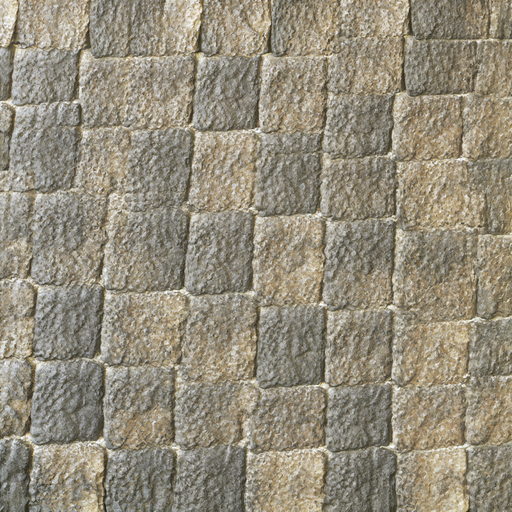} %
    \end{subfigure}\\

    \begin{subfigure}[t]{0.12\linewidth}
    \includegraphics[width=\linewidth]{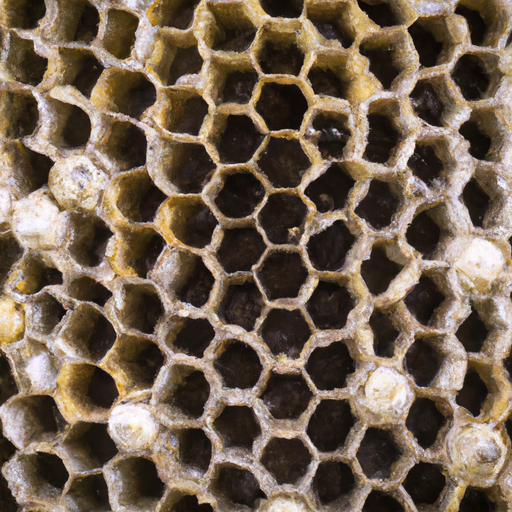} %
    \caption*{Input}
    \end{subfigure}
    \begin{subfigure}[t]{0.28\linewidth}
    \includegraphics[width=\linewidth]{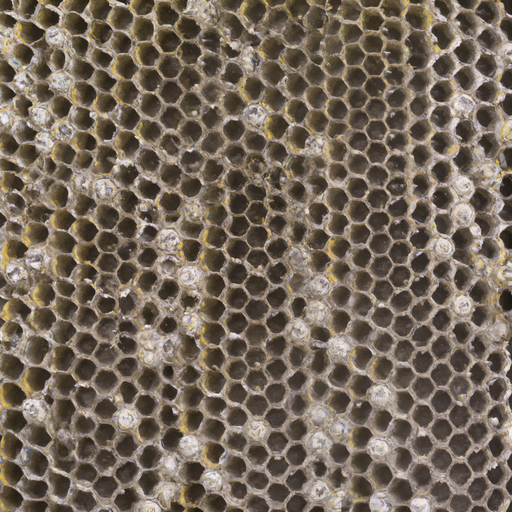} %
    \caption*{Fixed text encoder}
    \end{subfigure}
    \begin{subfigure}[t]{0.28\linewidth}
    \includegraphics[width=\linewidth]{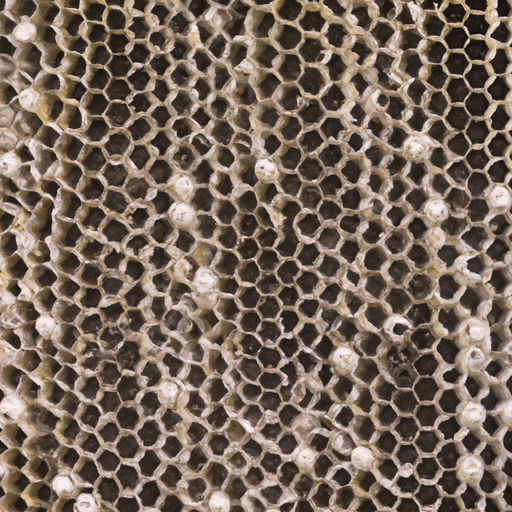} %
    \caption*{Fixed crops}
    \end{subfigure}
    \begin{subfigure}[t]{0.28\linewidth}
    \includegraphics[width=\linewidth]{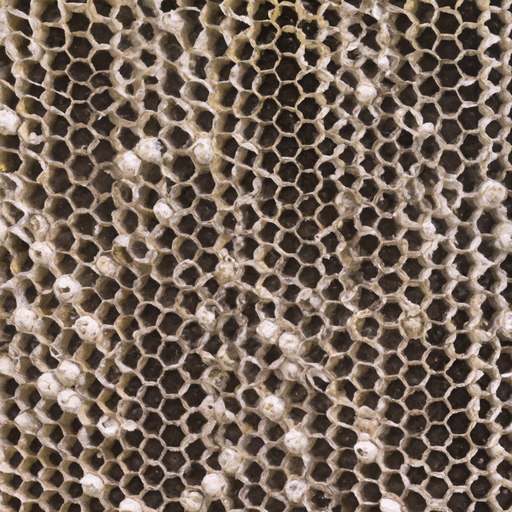} %
    \caption*{Ours}
    \end{subfigure}

    \vspace{-5pt}

    \caption{\textbf{Ablation studies.} 
    Fixing the text encoder introduces color drift due to inherited semantic priors.
    Conversely, using fixed crops at test time maintains the same image quality but extends runtime from 6 to 50 minutes versus random crops.
    }
    \label{fig:ablation}

    \vspace{-30pt}
    \end{center}
\end{figure}

\subsection{Texture Synthesis}
\label{sec:multidiffison}

A na\"{\i}ve approach to generating a high-resolution texture image is to directly denoise a large latent code map $\bz_{t_1} \sim \mathcal{N}(\bzero, \bI)$ using the fine-tuned diffusion model.
However, due to the computational limitations imposed by the model's size, it is not feasible to denoise such high-resolution latent code, \eg $2048 \times 2048$, on a single GPU. 
To address this constraint, we adopt a progressive denoising strategy inspired by MultiDiffusion~\cite{bar2023multidiffusion}.

Instead of denoising the entire noisy latent map at once, we take a patch-by-patch decoding approach similar to traditional texture synthesis methods. 
We extract a set of random crops $F_i$, that crops the latent map to a constant resolution, specifically $96 \times 96$.
Since the decoder of Stable Diffusion upsamples the latent map by a factor of $8$, $F_i$ would translate to an image of size $768 \times 768$.
At each denoising time step $t$, we employ a score aggregation strategy. We begin by denoising the random crops in the latent space:
\begin{equation}
    \hat{\bx}_{0_i}^t = \hat\bx_{\theta_\text{ref}}(F_i (\bz_t), \bc)
\end{equation}

where $\bz_t$ is the noisy latent map at time step $t$ and $\hat{\bx}_{0_i}^t$ is the denoised patch of $F_i (\bz_t)$.
We then combine the denoised patch $\hat{\bx}_{0_i}^t$ to obtain the large $\bx$-prediction map $\hat{\bx}_0^t$. 
Following~\cite{bar2023multidiffusion}, we formulate the combination of $\hat{\bx}_{0_i}^t$ as an optimization problem:

\begin{equation}
    \hat{\bx}_0^t = \argmin_\bx \sum_{i=1}^n \| F_i (\bx) - \hat\bx_{\theta_\text{ref}}(F_i (\bz_t), \bc) \|^2_2
    \label{eq:md_loss}
\end{equation}

This optimization problem aims to reconcile all denoised samples $\hat\bx_{\theta_\text{ref}}(F_i (\bz_t), \bc)$ obtained from different random crops. By minimizing the loss in Eq.~\ref{eq:md_loss}, we ensure that denoised samples from different regions are combined as consistently as possible. Eq.~\ref{eq:md_loss} is a quadratic Least-Squares and has a closed-form solution: each pixel of the minimizer $\hat{\bx}_0^t$ is an average of all denoised sample updates.

Intuitively, we synthesize a high-resolution texture in a score aggregation manner. At every time step $t$. we first divide the large noisy latent map $\bz_t$ into small patches and denoise each patch using the fine-tuned diffusion model. These denoised patches are then aggregated by averaging. We found that using random crops achieves the same image quality as using fixed crops~\cite{bar2023multidiffusion}, while significantly speeding up the inference time, by a factor of $10$.
By denoising only small patches, we also overcome memory constraints. Furthermore, our algorithm scales well with resolution since the patch size remains fixed. This allows our method to generate arbitrarily large high-quality textures based on the reference texture image.

\begin{figure*}[!htbp]
    \begin{center}

    \begin{subfigure}[t]{0.49\linewidth}
        \includegraphics[width=0.32\linewidth]{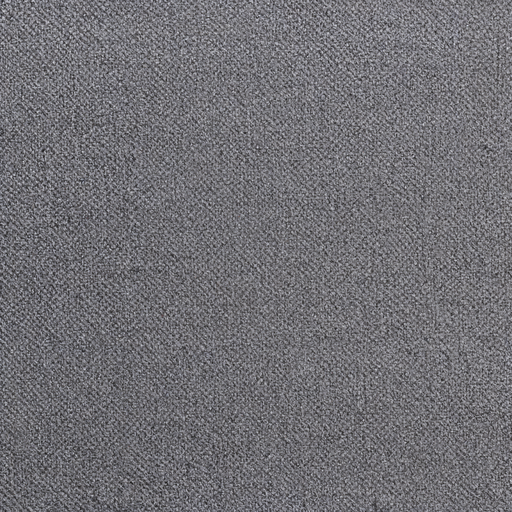} %
        \includegraphics[width=0.32\linewidth]{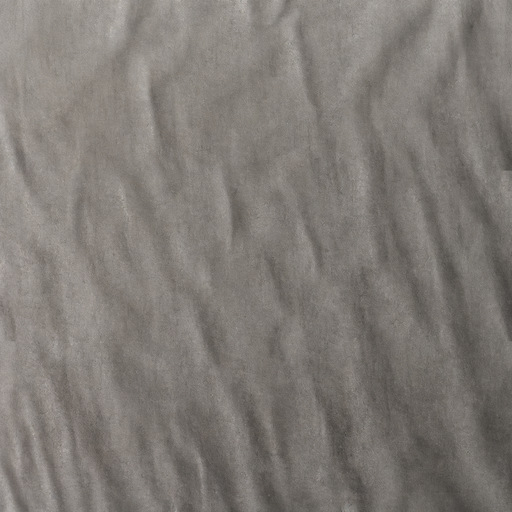} %
        \includegraphics[width=0.32\linewidth]{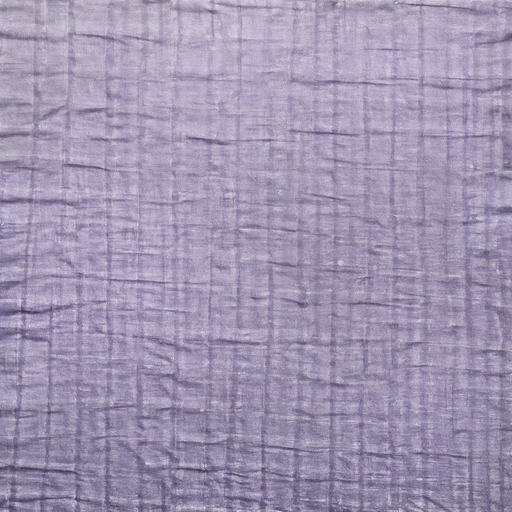} %
        \caption*{\textit{orthographic surface texture of cloth}}
    \end{subfigure} 
    \begin{subfigure}[t]{0.49\linewidth}
        \includegraphics[width=0.32\linewidth]{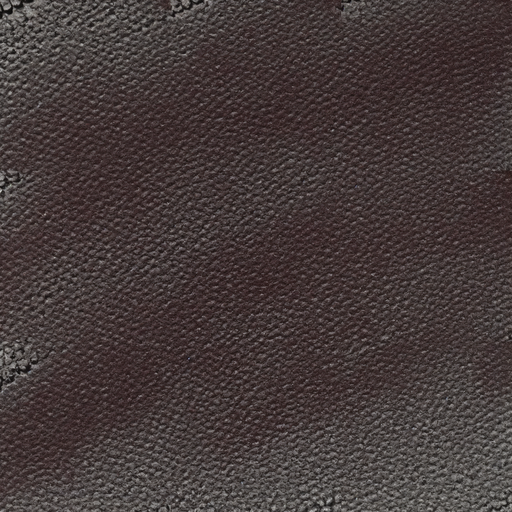} %
        \includegraphics[width=0.32\linewidth]{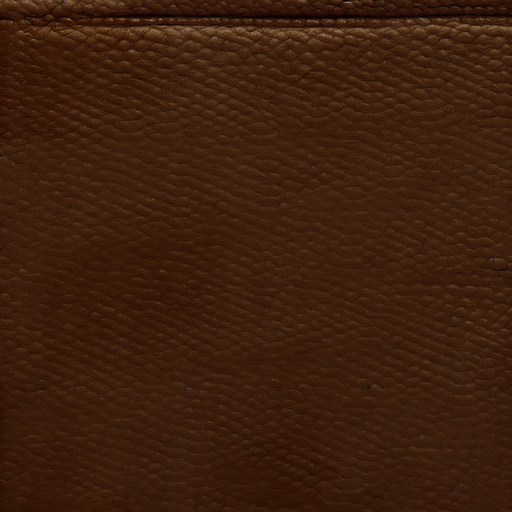} %
        \includegraphics[width=0.32\linewidth]{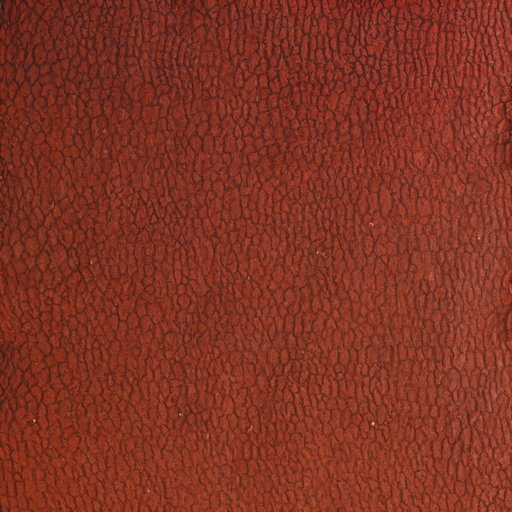} %
        \caption*{\textit{orthographic surface texture of leather}}
    \end{subfigure} \\

    \begin{subfigure}[t]{0.49\linewidth}
        \includegraphics[width=0.32\linewidth]{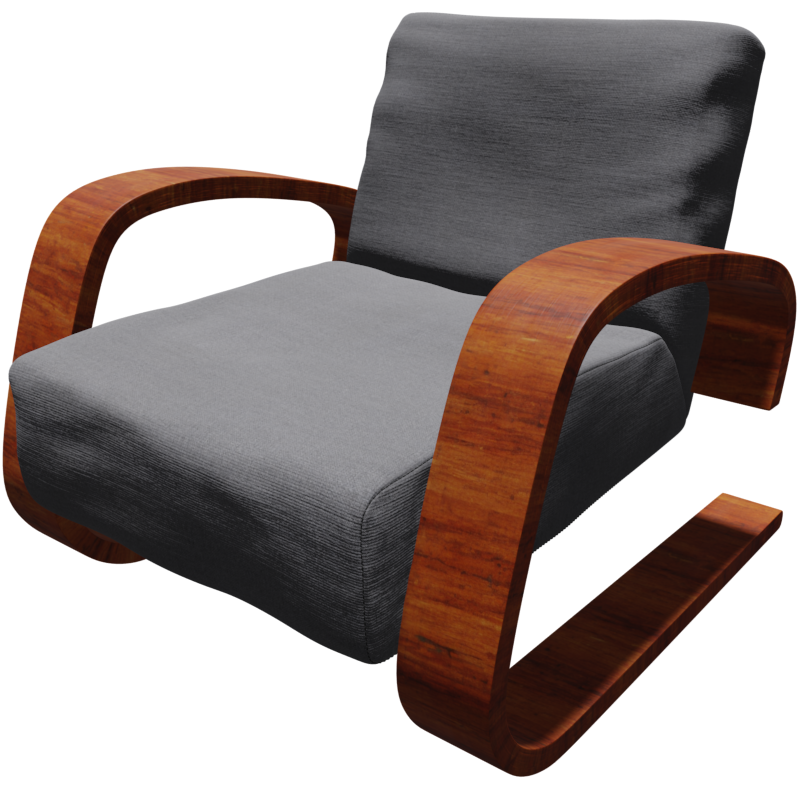} %
        \includegraphics[width=0.32\linewidth]{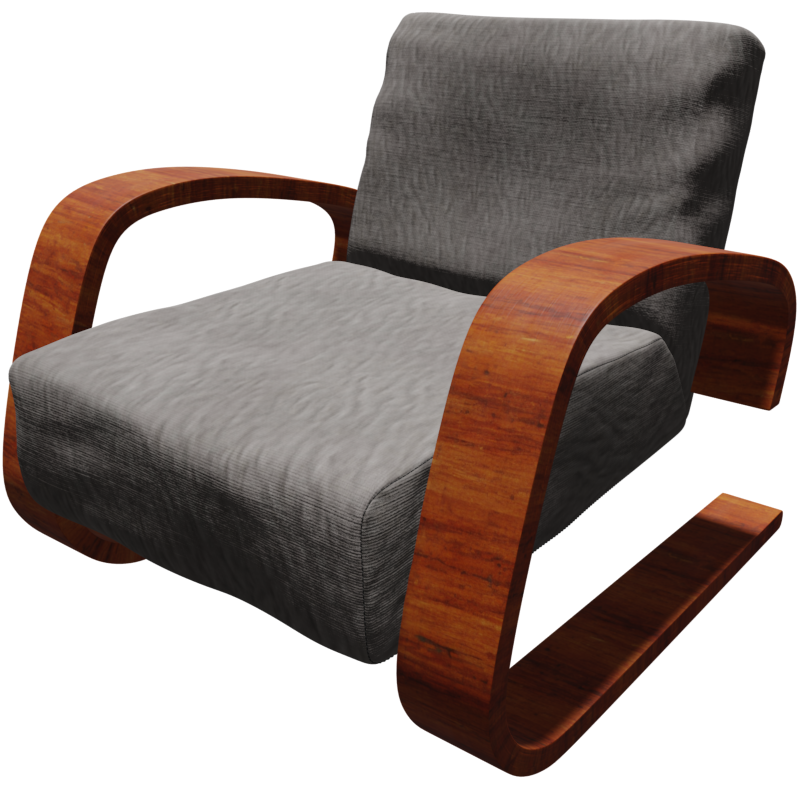} %
        \includegraphics[width=0.32\linewidth]{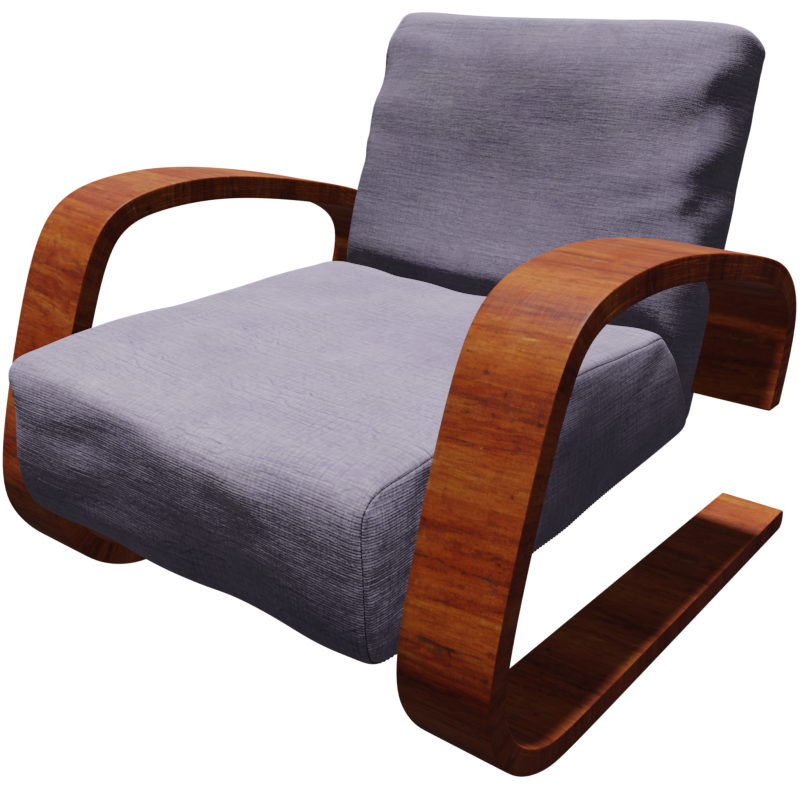} %
    \end{subfigure} 
    \begin{subfigure}[t]{0.49\linewidth}
        \includegraphics[width=0.32\linewidth]{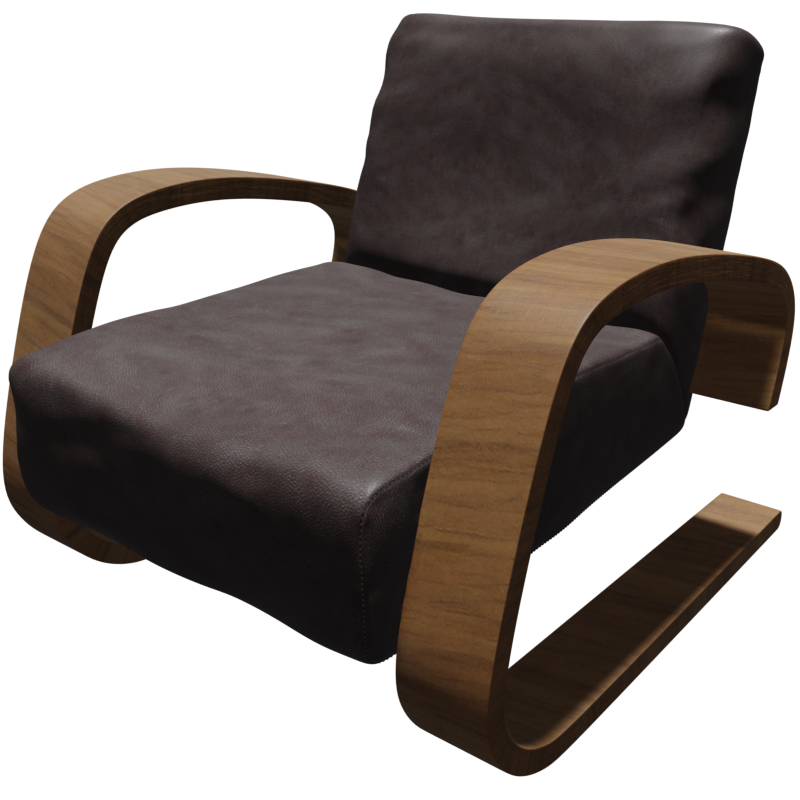} %
        \includegraphics[width=0.32\linewidth]{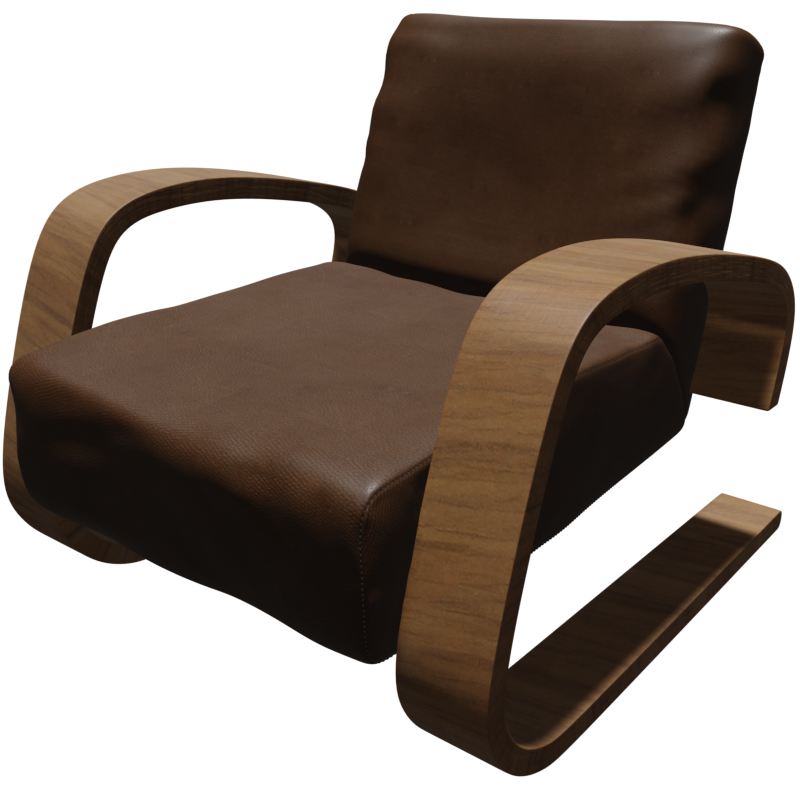} %
        \includegraphics[width=0.32\linewidth]{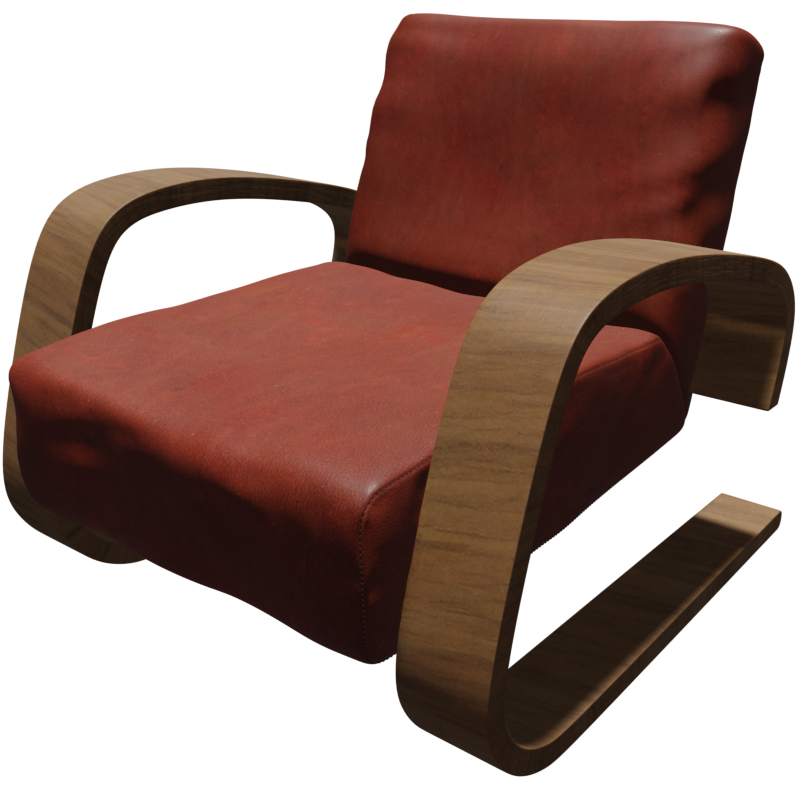} %
    \end{subfigure}

    \caption{\textbf{Results of synthesized textures and renderings}. We demonstrate high-resolution textures generated by \model, along with photo-realistic renderings of an armchair utilizing the textures (including the wood armrest). \model is capable of generating textures with different variations when provided with the same text prompt. These textures can be seamlessly incorporated into 3D rendering pipelines, resulting in nearly infinite material choices for assets in a 3D shape collection. We use the default UV mapping from the CAD model to warp the texture onto the 3D model.
    }
    \label{fig:resutls_ours}

    \end{center}
\end{figure*}
\begin{figure*}[!htbp]
    \centering

    \begin{subfigure}[t]{0.16\linewidth}
    \includegraphics[width=\linewidth]{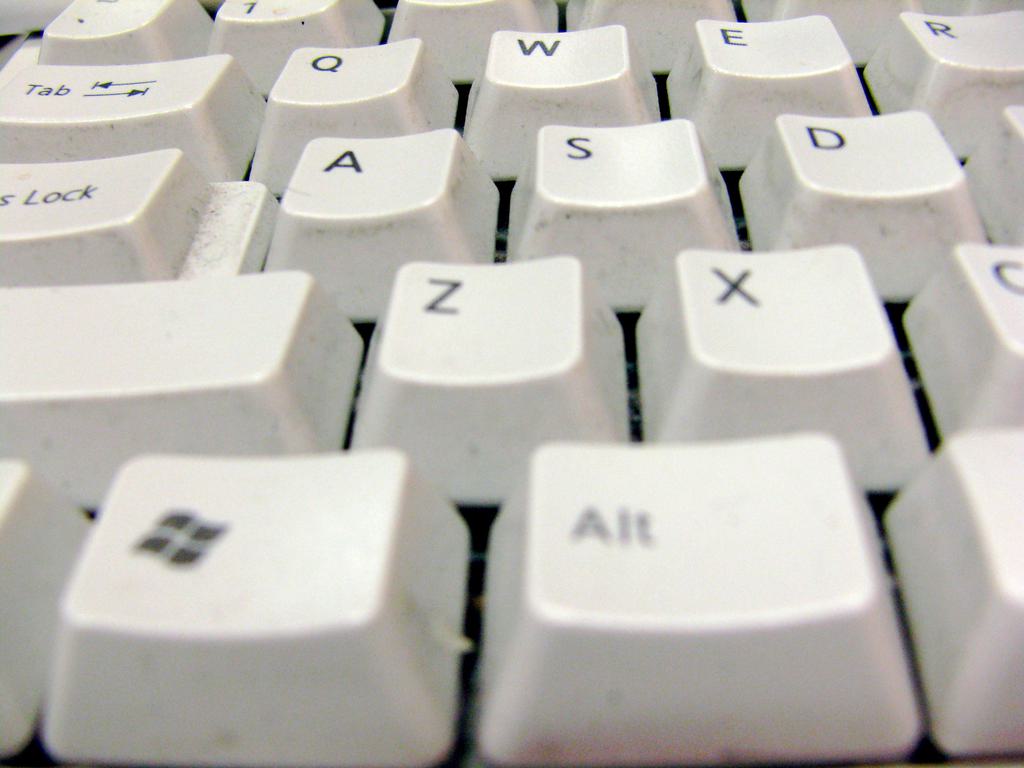} %
    \end{subfigure}
    \begin{subfigure}[t]{0.16\linewidth}
    \includegraphics[width=\linewidth]{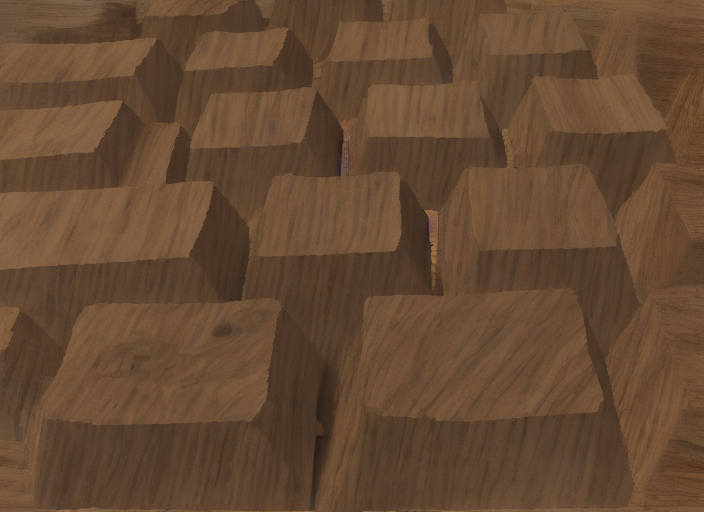} %
    \end{subfigure}
    \begin{subfigure}[t]{0.16\linewidth}
    \includegraphics[width=\linewidth]{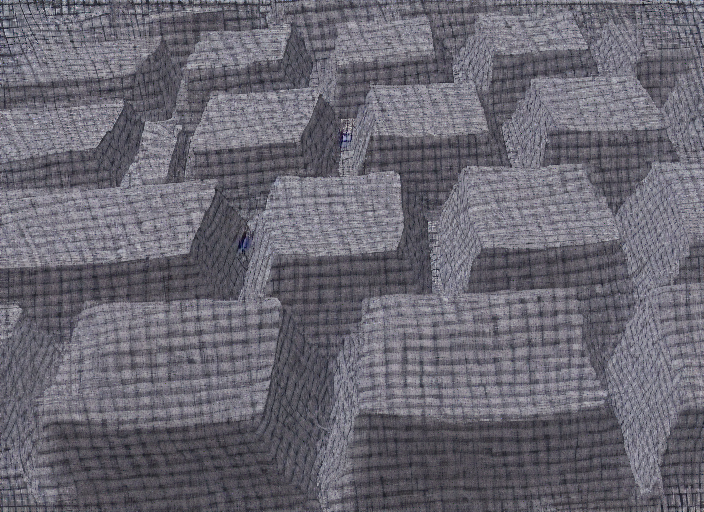} %
    \end{subfigure}
    \begin{subfigure}[t]{0.16\linewidth}
    \includegraphics[width=\linewidth]{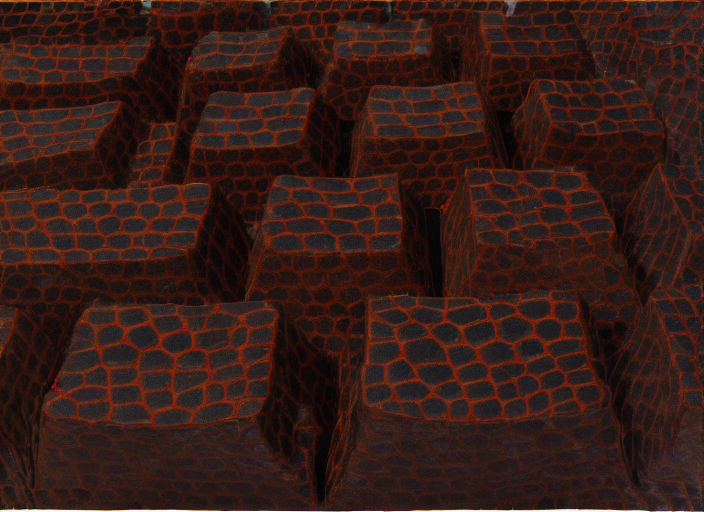} %
    \end{subfigure}
    \begin{subfigure}[t]{0.16\linewidth}
    \includegraphics[width=\linewidth]{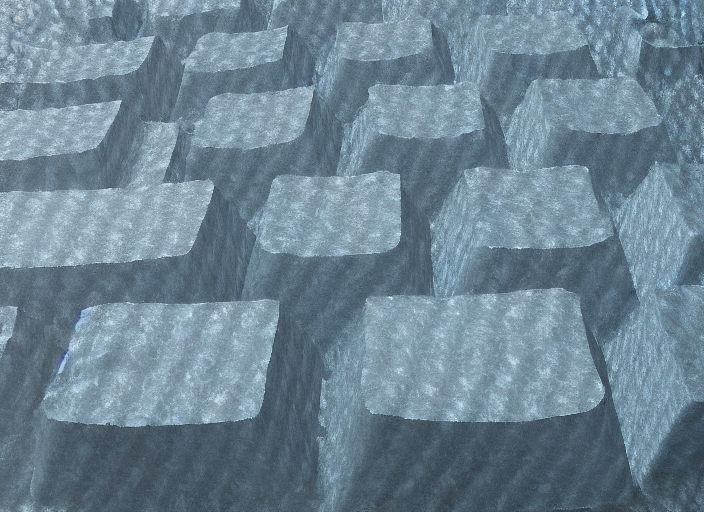} %
    \end{subfigure}
    \begin{subfigure}[t]{0.16\linewidth}
    \includegraphics[width=\linewidth]{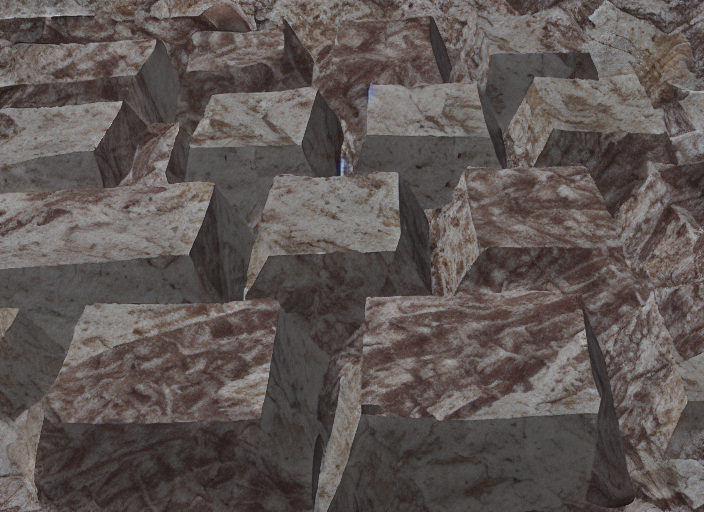} %
    \end{subfigure} \\

    \begin{subfigure}[t]{0.16\linewidth}
    \includegraphics[width=\linewidth]{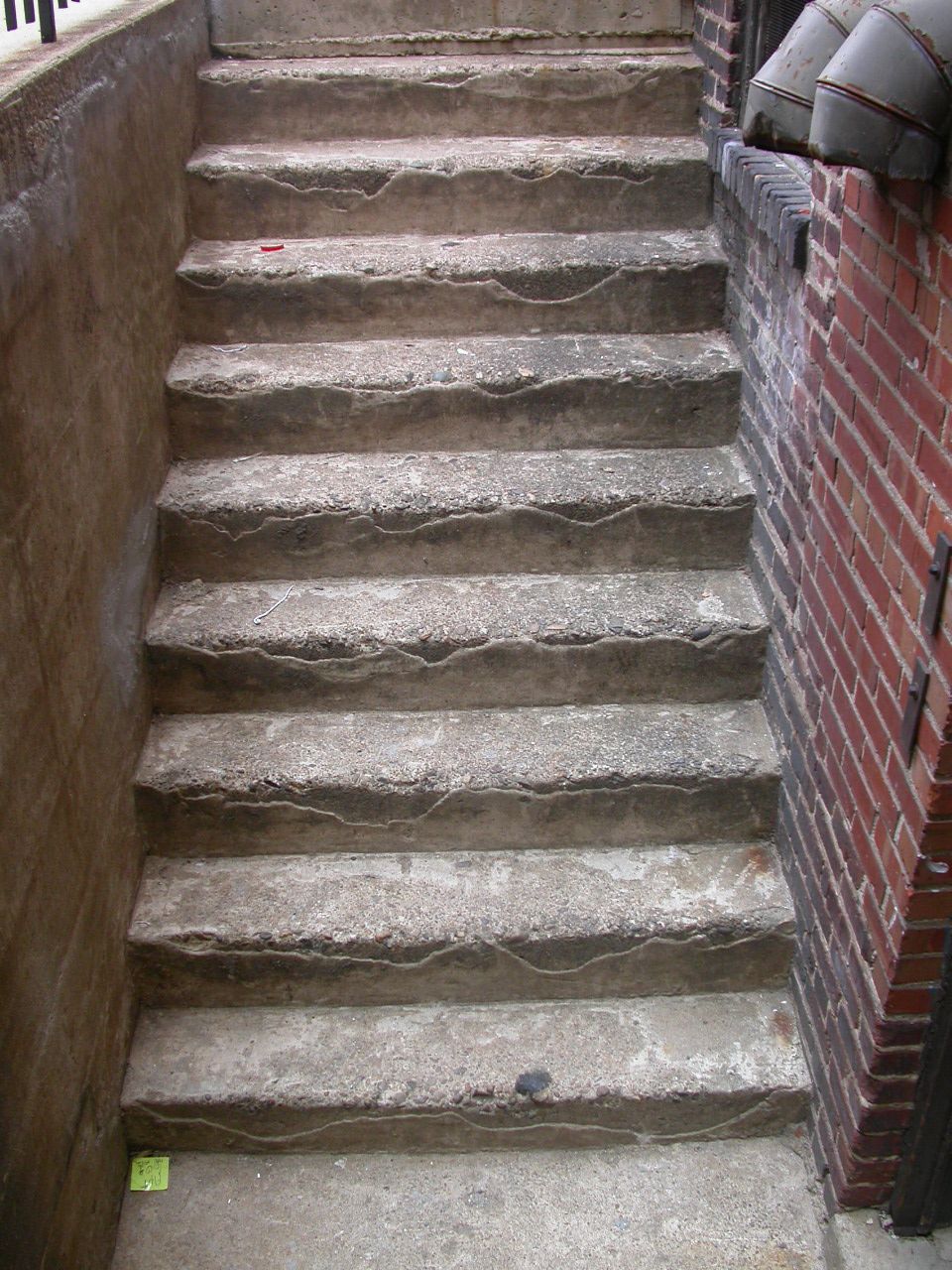} %
    \end{subfigure}
    \begin{subfigure}[t]{0.16\linewidth}
    \includegraphics[width=\linewidth]{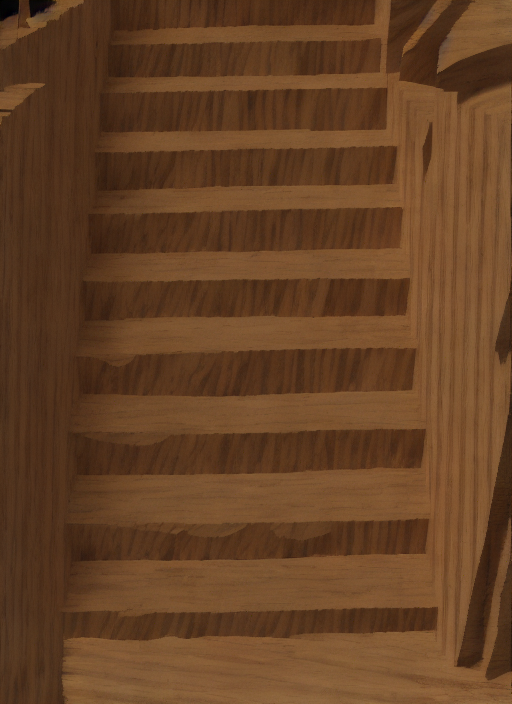} %
    \end{subfigure}
    \begin{subfigure}[t]{0.16\linewidth}
    \includegraphics[width=\linewidth]{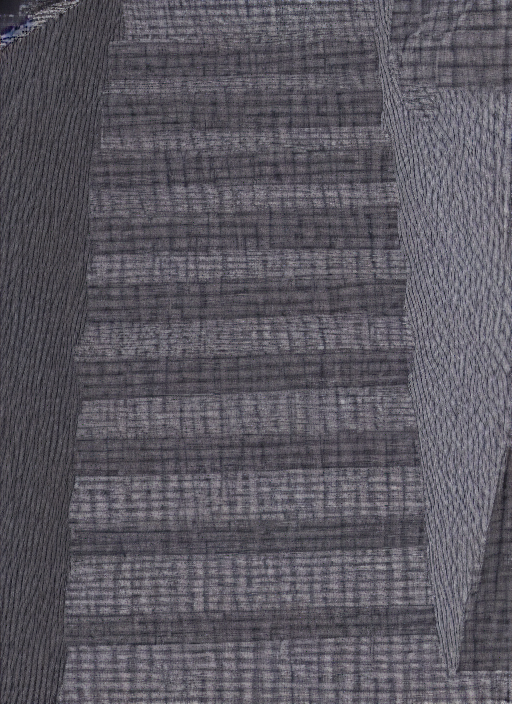} %
    \end{subfigure}
    \begin{subfigure}[t]{0.16\linewidth}
    \includegraphics[width=\linewidth]{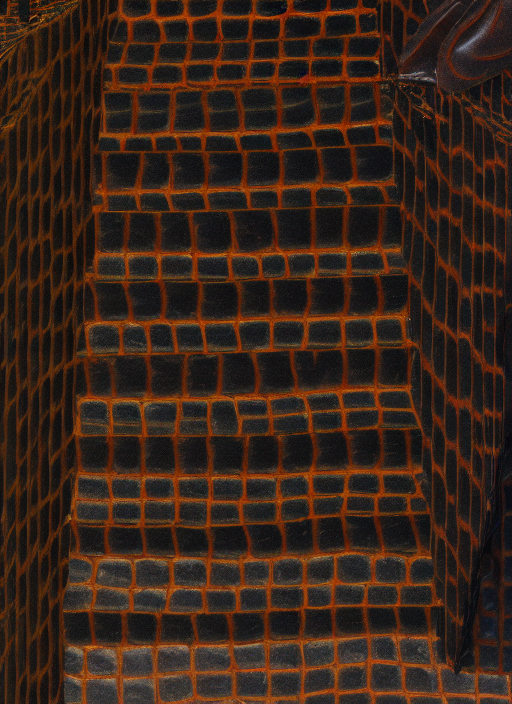} %
    \end{subfigure}
    \begin{subfigure}[t]{0.16\linewidth}
    \includegraphics[width=\linewidth]{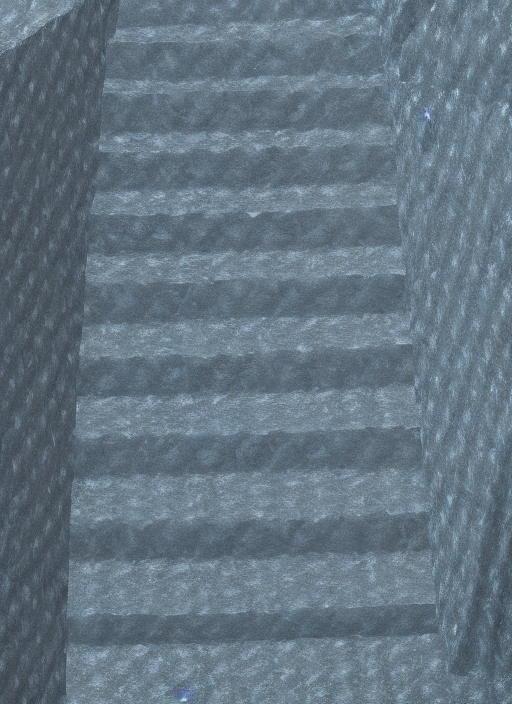} %
    \end{subfigure}
    \begin{subfigure}[t]{0.16\linewidth}
    \includegraphics[width=\linewidth]{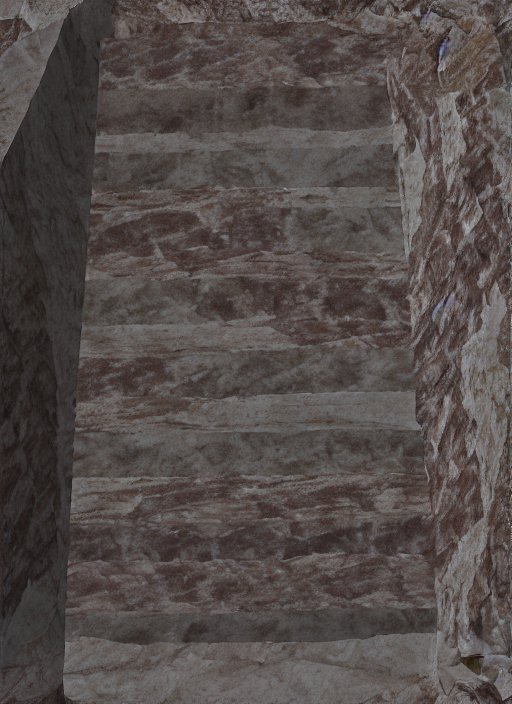} %
    \end{subfigure} \\
    
    \begin{subfigure}[t]{0.16\linewidth}
    \includegraphics[width=\linewidth]{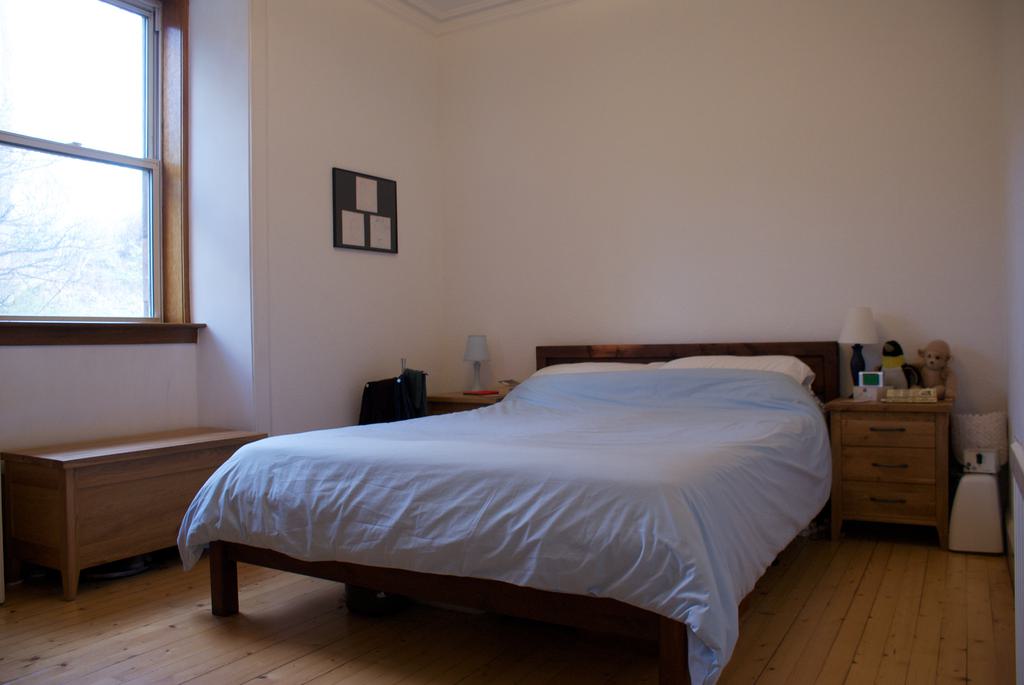} %
    \end{subfigure}
    \begin{subfigure}[t]{0.16\linewidth}
    \includegraphics[width=\linewidth]{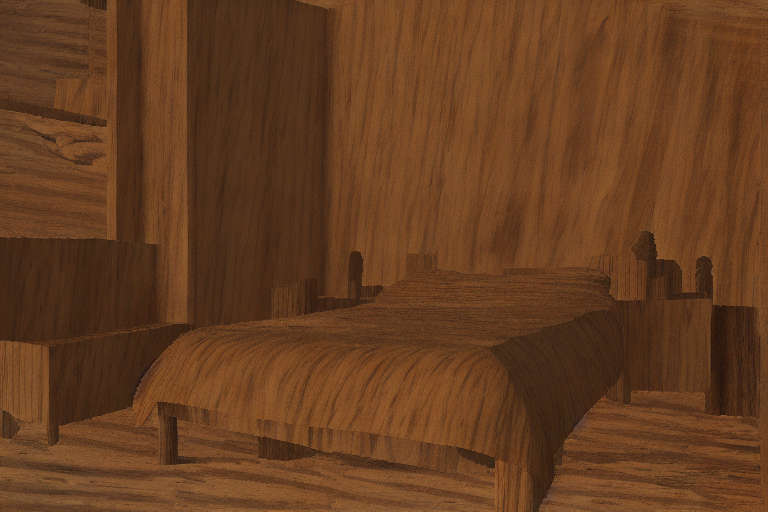} %
    \end{subfigure}
    \begin{subfigure}[t]{0.16\linewidth}
    \includegraphics[width=\linewidth]{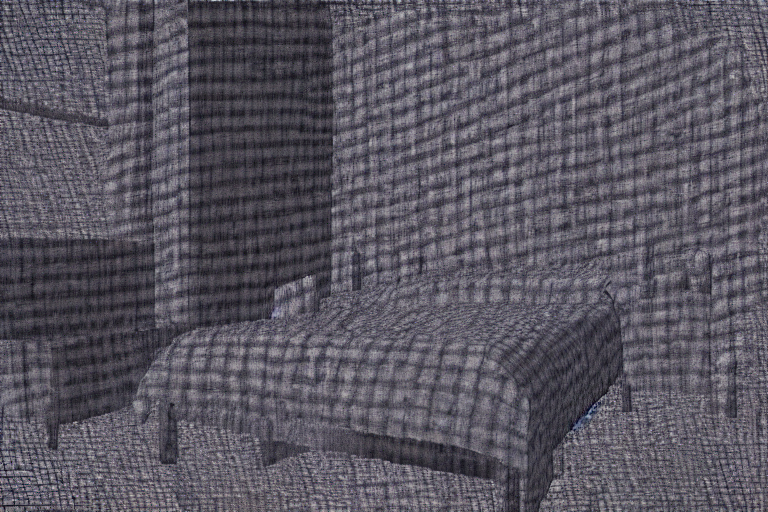} %
    \end{subfigure}
    \begin{subfigure}[t]{0.16\linewidth}
    \includegraphics[width=\linewidth]{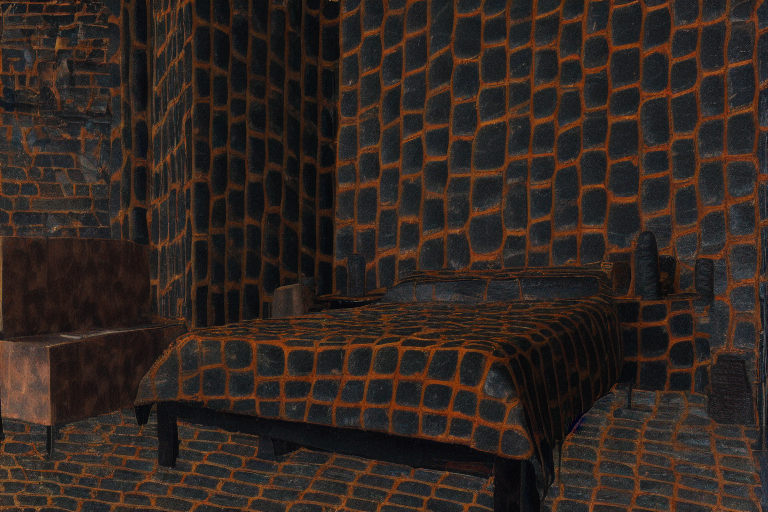} %
    \end{subfigure}
    \begin{subfigure}[t]{0.16\linewidth}
    \includegraphics[width=\linewidth]{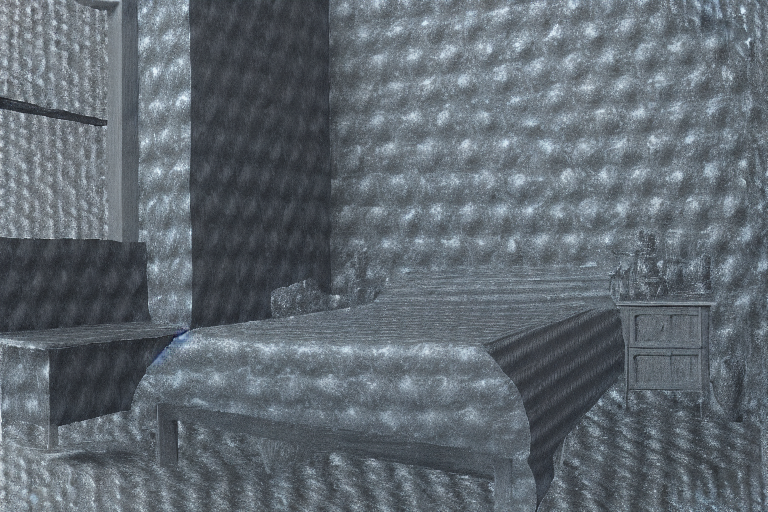} %
    \end{subfigure}
    \begin{subfigure}[t]{0.16\linewidth}
    \includegraphics[width=\linewidth]{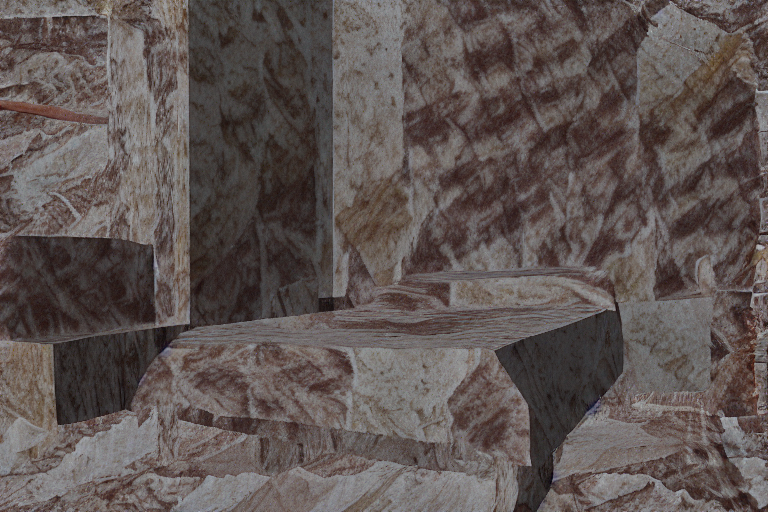} %
    \end{subfigure} \\

    \begin{subfigure}[t]{0.16\linewidth}
    \includegraphics[width=\linewidth]{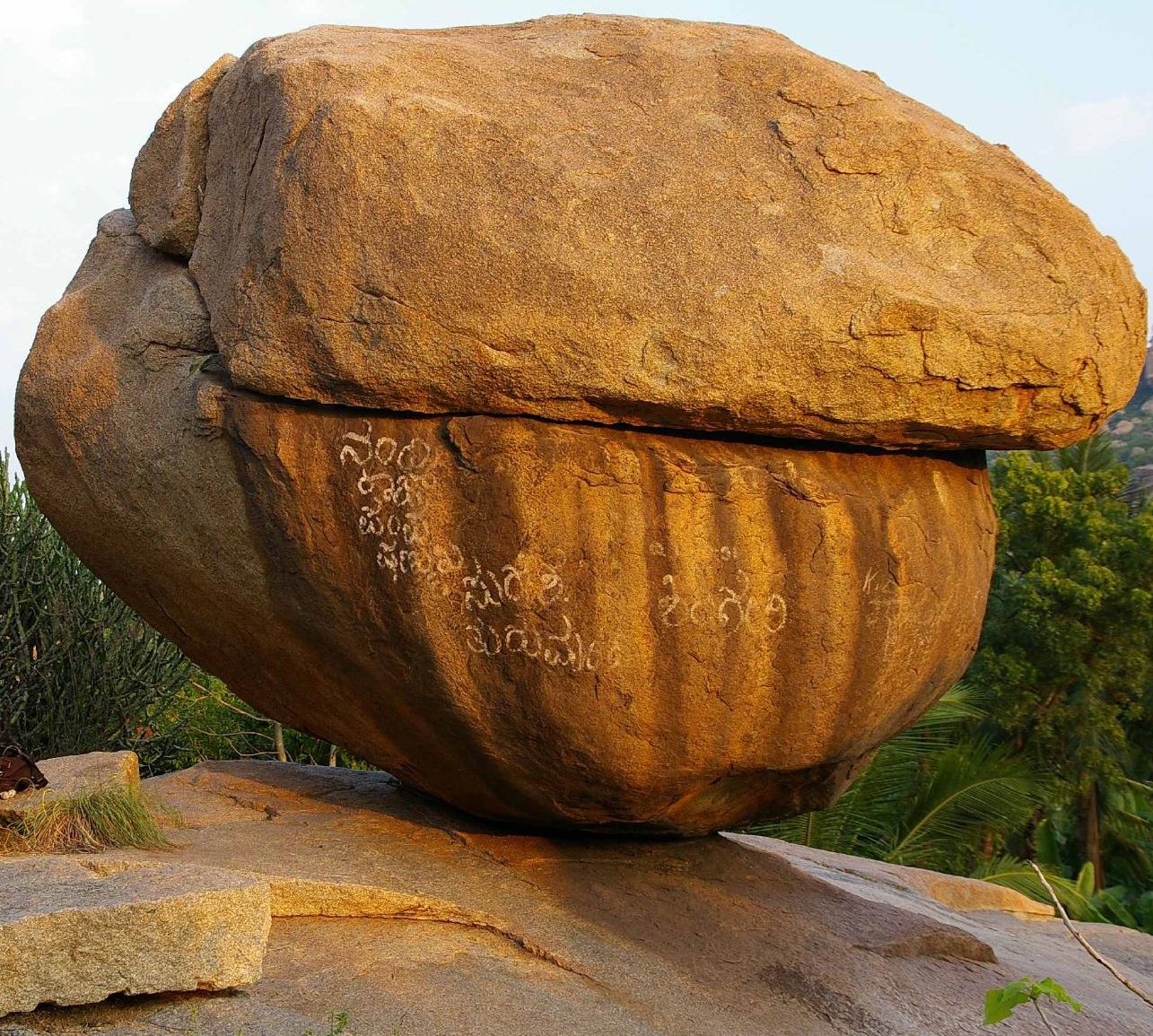} %
    \end{subfigure}
    \begin{subfigure}[t]{0.16\linewidth}
    \includegraphics[width=\linewidth]{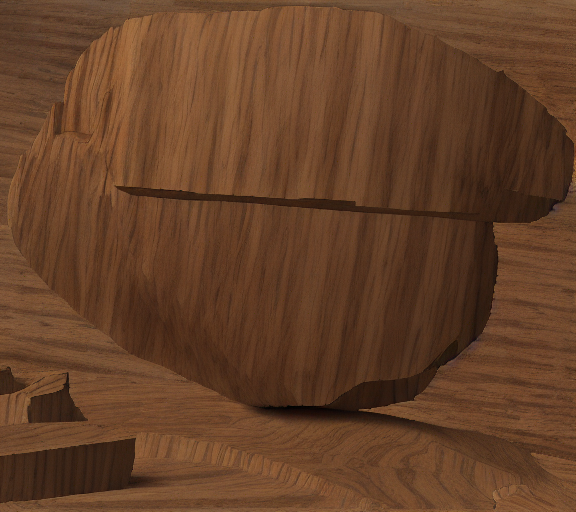} %
    \end{subfigure}
    \begin{subfigure}[t]{0.16\linewidth}
    \includegraphics[width=\linewidth]{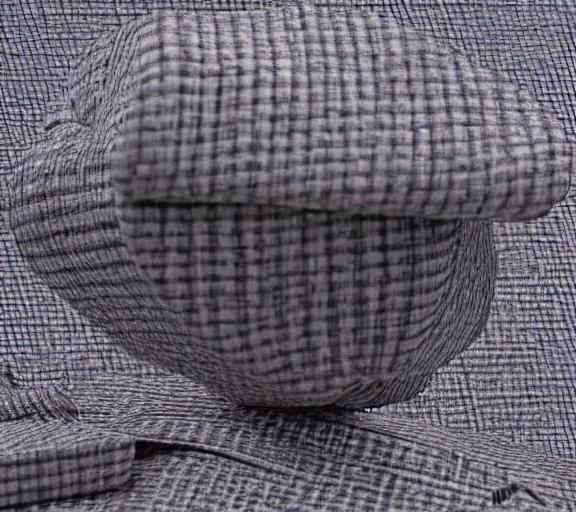} %
    \end{subfigure}
    \begin{subfigure}[t]{0.16\linewidth}
    \includegraphics[width=\linewidth]{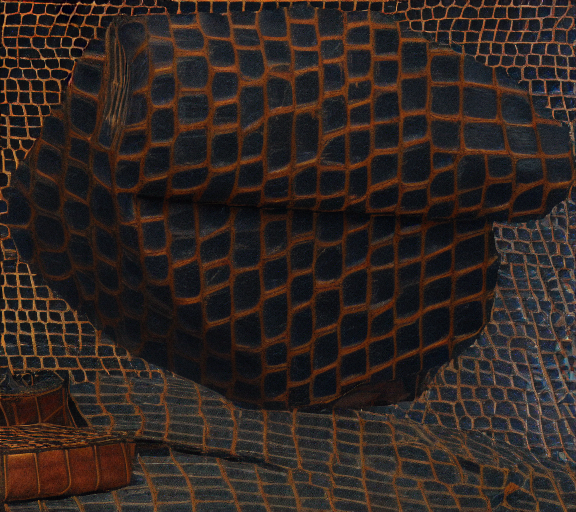} %
    \end{subfigure}
    \begin{subfigure}[t]{0.16\linewidth}
    \includegraphics[width=\linewidth]{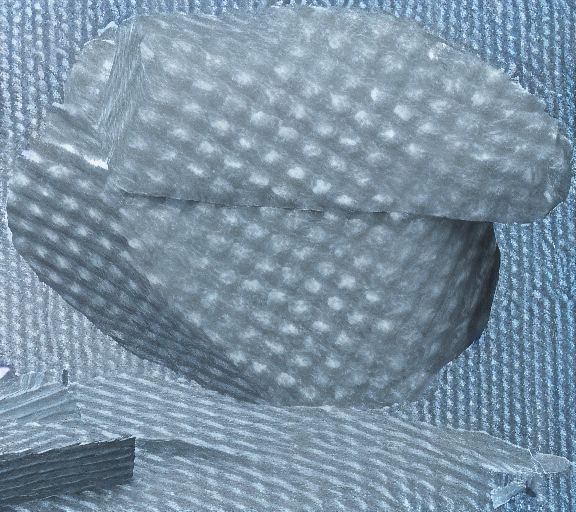} %
    \end{subfigure}
    \begin{subfigure}[t]{0.16\linewidth}
    \includegraphics[width=\linewidth]{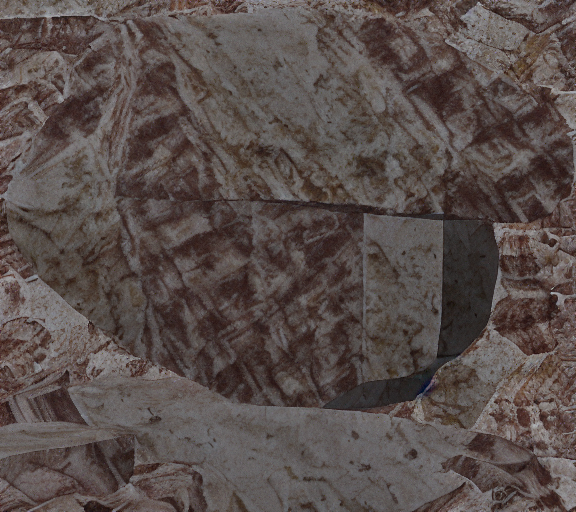} %
    \end{subfigure} \\ 

    \begin{subfigure}[t]{0.16\linewidth}
    \includegraphics[width=\linewidth]{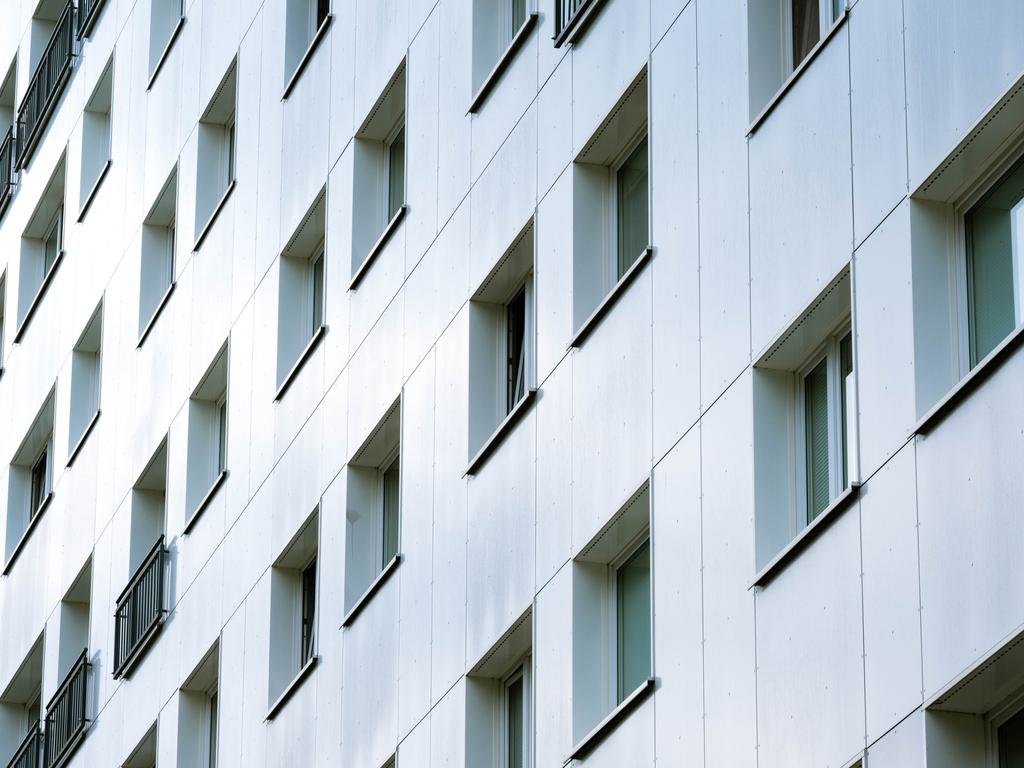} %
    \caption{Input}
    \end{subfigure}
    \begin{subfigure}[t]{0.16\linewidth}
    \includegraphics[width=\linewidth]{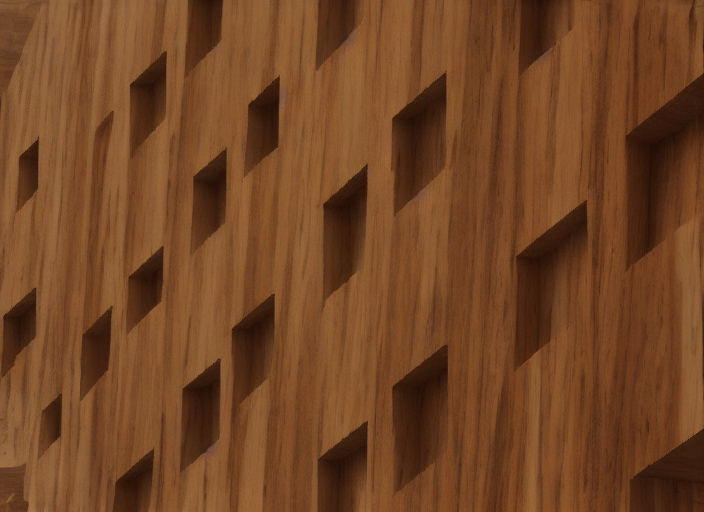} %
    \caption{\textit{Wood}}
    \end{subfigure}
    \begin{subfigure}[t]{0.16\linewidth}
    \includegraphics[width=\linewidth]{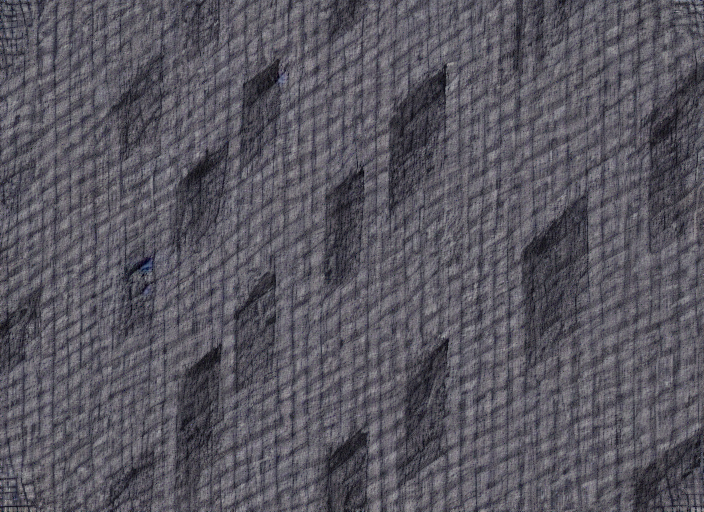} %
    \caption{\textit{Fabric}}
    \end{subfigure}
    \begin{subfigure}[t]{0.16\linewidth}
    \includegraphics[width=\linewidth]{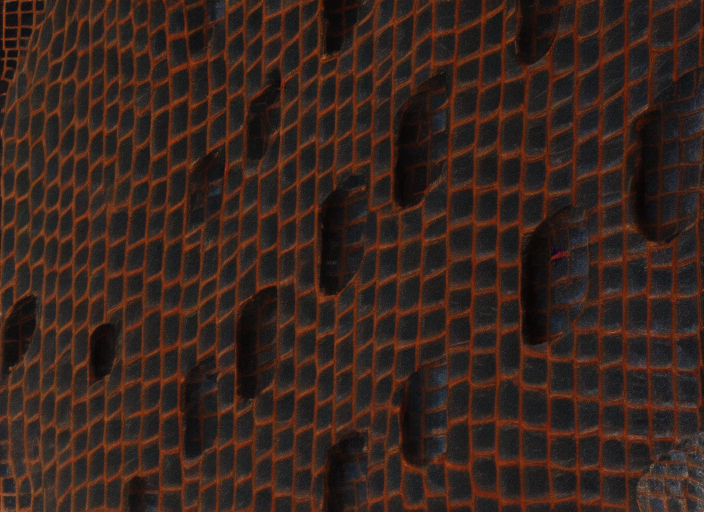} %
    \caption{\textit{Alligator skin}}
    \end{subfigure}
    \begin{subfigure}[t]{0.16\linewidth}
    \includegraphics[width=\linewidth]{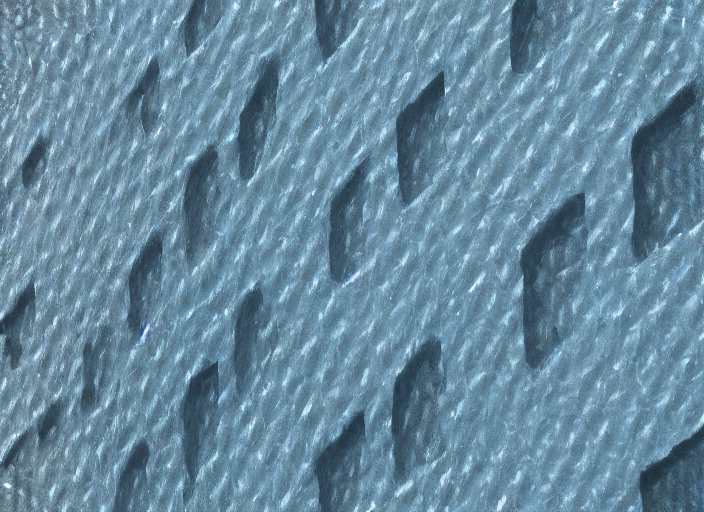} %
    \caption{\textit{Rubber floor mat}}
    \end{subfigure}
    \begin{subfigure}[t]{0.16\linewidth}
    \includegraphics[width=\linewidth]{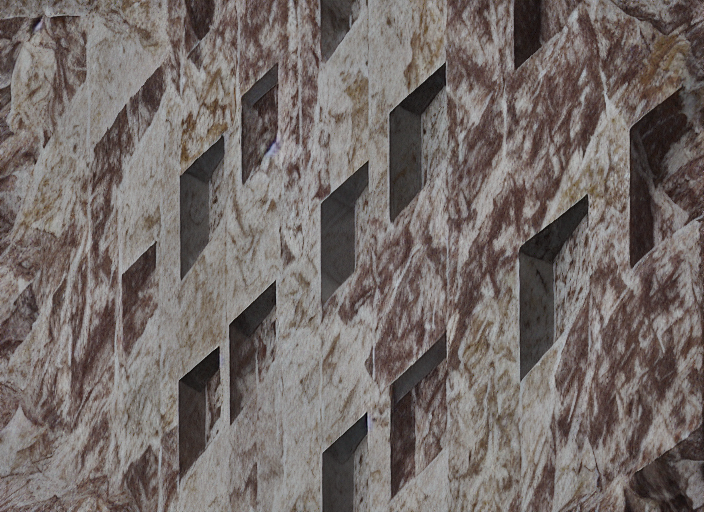} %
    \caption{\textit{Granite}}
    \end{subfigure}

    \caption{\textbf{Results of texture transfer.} We present an application of texture transfer facilitated by \model. We leverage ControlNet \cite{zhang2023adding} to transfer our generated textures to new surfaces. To achieve this, we first used MiDaS~\cite{Ranftl2022} to estimate the depth map from the input image. The depth map then served as the conditioning signal to control the texture's application to the surface observed in the real image. Re-textured results preserve the statistics of the input texture, while sharing consistent shading and shape with the input image.}

    \label{fig:retex}
\end{figure*}

\section{Evaluation}
In this section, we perform ablation studies to evaluate the importance of each model component, and compare \model against both traditional and deep learning-based state-of-the-art in the tasks of texture synthesis. 

\subsection{Ablation Studies}
We perform qualitative ablation on \model to evaluate the importance of tuning a text encoder and the impact of using random (vs. fixed) crops.

\textit{Fixed vs. trainable text encoder.}
We conducted a comparison between fine-tuning diffusion models with a fixed text encoder and a trainable text encoder. Fixing the text encoder helps maintain semantic priors, but it also introduces a false signal to the image synthesis module through the unique identifier.
As illustrated in Fig.~\ref{fig:ablation}, fine-tuning the diffusion model while also training the text encoder preserves the most image priors and results in reduced color drift.

\textit{Fixed vs. random crops.}
At inference time, we employ a set of random crops $F_i$ to denoise the random patches in the large latent map. In contrast, the original MultiDiffusion~\cite{bar2023multidiffusion} utilizes a set of fixed crops $F_i$. As shown in Fig.~\ref{fig:ablation}, using random crops achieves comparable image quality while significantly improving the runtime by a factor of $10$.

\subsection{Baseline Comparisons}
We conducted qualitative comparisons of \model with a variety of texture synthesis baselines, covering both traditional~\cite{efros2001image,kaspar2015self} and deep learning-based methods~\cite{bergmann2017learning,zhou2018non}. 
The field of quantitative comparisons in texture synthesis continues to be an active area of research~\cite{lin2006quantitative}.
Standard quantitative metrics, \eg, Gram matrix computation or FID, are not optimal choices for evaluation as they are used as objective functions by certain methods.
Instead, we conducted a human study to quantitatively assess the realism of synthesized textures.

\textbf{Image Quilting.}
As an early texture synthesis method, Image Quilting~\cite{efros2001image}  is a patch-based approach. It involves tiling an example patch into a grid and utilizing dynamic programming to determine an optimal path for cutting through the overlapping regions, and ideally results in a seamless tile composed of the same patch. However, when applied to high-resolution textures, the complexity of solving the dynamic programming increases, leading to the generation of repetitive patterns. 
As shown in Fig.~\ref{fig:qualitative}, image quilting tends to produce noticeable repetitive patterns when working with real-world high-resolution texture patches.

\textbf{Self Tuning Texture Optimization.}
While non-parametric method~\cite{efros2001image} struggled with real-world high-resolution textures, 
Self Tuning Texture Optimization (STTO)~\cite{kaspar2015self} is a fully automatic self-tuning texture sytnthesis method that extends Texture Optimization~\cite{kwatra2003graphcut,darabi2012image} to handle textures with large-scale structures, repetitions, and near-regular structures. STTO is capable of self-tuning its various parameters, thereby eliminating the need for manual adjustment of parameters on a case-by-case basis.
As shown in Fig.~\ref{fig:qualitative}, STTO tends to produce results with small broken structures. It also fails with textures that contain large but unpronounced features (\eg fabric and wood in Fig.~\ref{fig:qualitative}), due to the contour detector's inability to detect reliable edges.

\textbf{Periodic Spatial GAN.}
Periodic Spatial GAN (PSGAN)~\cite{bergmann2017learning} is a texture synthesis method based on Generative Adversarial Networks (GANs). PSGAN extends the DCGAN~\cite{radford2015unsupervised} network structure by incorporating a spatially periodic signal generator to learn the statistical properties of the given texture image.
As illustrated in Fig.~\ref{fig:qualitative}, PSGAN fails to capture the frequency of the texture image. This limitation arises from the periodic signal generator's preference for high-frequency textures and the constrained receptive field size.

\textbf{Non-Stationary Texture Synthesis.}
Non-stationary texture synthesis (NSTS)~\cite{zhou2018non} is the state-of-the-art method for texture synthesis. It trains a GAN per input texture image to double the spatial extent of texture blocks extracted from the input exemplar texture. 
Once trained, the fully convolutional generator is capable of expanding the size of the input texture image. The network structure is based on CycleGAN~\cite{zhu2017unpaired}.
As shown in Fig.~\ref{fig:qualitative}, NSTS generates visually pleasing textures but fails to accurately follow the distribution of the input texture. This limitation stems from the network being trained solely on small patches ($256 \times 256$) of the example texture, resulting in overfitting to low-frequency details.

\textbf{Human study.}
To quantify the realism of synthesized textures, we conducted a human study via Amazon Mechanical Turk (AMT). This study involved a set of 39 exemplar textures.
For each exemplar texture, we presented five synthesized textures from various methods to three human subjects, and asked them to choose the one that most closely resembles the exemplar texture. We also provided instructions to specifically note artifacts such as visible seams, noticeable repetition patterns, color drifts, loss of sharpness, irregular patterns, and inconsistent global structures.
Our method was chosen as the best 45\% of the time, compared to NSTS at 22\%, Image Quilting at 15\%, STTO at 13\%, and PSGAN at 5\%, demonstrating a clear advantage in the task of texture synthesis.

\section{Applications}
In this section, we present more results of \model, and its applications in synthetic rendering (Sec.~\ref{sec:rendering}) and retexturing natural images (Sec.~\ref{sec:retex}).

\subsection{Texture Generation and Application to 3D Models}
\label{sec:rendering}
We showcase sample textures generated by \model in Fig.~\ref{fig:resutls_ours}. \model takes in the text prompt, and leverages DALL-E~2~\cite{ramesh2022hierarchical} to synthesize a reference texture image of size $1024 \times 1024$. We then fine-tune a Stable Diffusion v2~\cite{rombach2021highresolution} checkpoint for 1000 iterations, which takes 10 minutes on a single NVIDIA A100 GPU. At inference time, it takes 6 minutes to generate a texture of size $2304 \times 2304$ or $1.5$ hours for a 85 MP texture in Fig~\ref{fig:teaser}. 
As shown in Fig.~\ref{fig:resutls_ours}, \model is able to synthesize textures of different frequencies, ranging from coarse fabric to dense leather grain. Moreover, it is capable of generating multiple distinct textures from the same text prompt, offering users a range of texture variations.
The generated texture images have direct applications in 3D rendering pipelines. As shown in Fig.~\ref{fig:resutls_ours}, these textures are applied to the same armchair, demonstrating \model's ability to provide high-quality, realistic appearance models for large-scale 3D shape collections.

\subsection{Texture Transfer}
\label{sec:retex}
Texture transfer is a process where a given texture is applied to another image, guided by various properties of the latter. Early work~\cite{efros2001image,hertzmann2001image} perform
texture transfer based on the brightness of the target image. Deep learning based methods~\cite{gatys2015neural,johnson2016perceptual,zhu2017unpaired} achieved texture transfer by aligning statistics of feature maps in the network via a style loss.
In our approach, we formulate texture transfer as an image synthesis task conditioned on scene depth.

We leverage ControlNet \cite{zhang2023adding} to transfer our generated textures to new surfaces. To achieve this, we first employ MiDaS~\cite{Ranftl2022} to estimate the depth map from the input image.
The depth map is then used as the conditional signal to model the texture's application to the surface. Following Zhang~\etal~\cite{zhang2023adding}, we incorporate the conditional depth map into a pretrained diffusion model using ControlNet. ControlNet takes a text prompt and an estimated depth map of the input image as inputs, and transfers the example texture onto the input image.

The ControlNet is fine-tuned on a minimal dataset constructed from texture-mapped primitives, and can generalize to nature images.
To construct the minimal dataset,
we use the generated texture of size $2304 \times 2304$ from \model to synthetically render a collection of image/depth pairs. 
For consistently shaded results, these images are rendered with the same texture applied to surfaces at various orientations, all under globally consistent shading.
The ControlNet is built upon a Stable Diffusion checkpoint. 
We use the same loss as in Eq.~\ref{eq:diffusion} to train both the ControlNet and the decoder part of the Stable Diffusion. More details are discussed in the supplemental material.

Fig.~\ref{fig:retex} showcases a collection of our texture transfer results. 
Our method successfully transfers the appearance of the example texture onto various surfaces.
We present results using the same texture applied across the entire scene to demonstrate that synthesized images have consistent shading and shape with the input image.
For future work, a segmentation model could be added to allow different textures on various regions of the scene.

\begin{figure}[!t]
    \begin{center}

    \begin{subfigure}[t]{0.38\linewidth}
    \includegraphics[width=\linewidth]{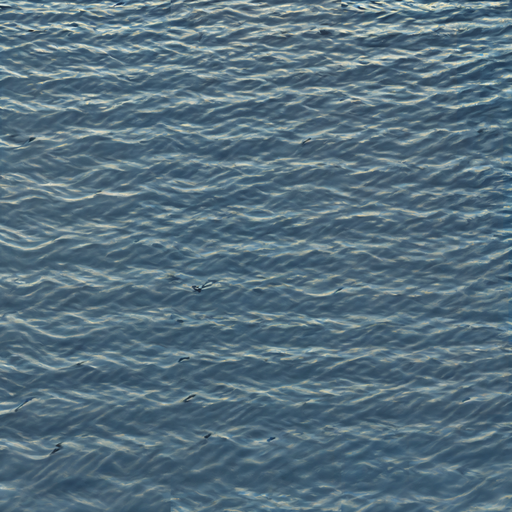} %
    \end{subfigure}
    \begin{subfigure}[b]{0.19\linewidth}
    \includegraphics[width=\linewidth]{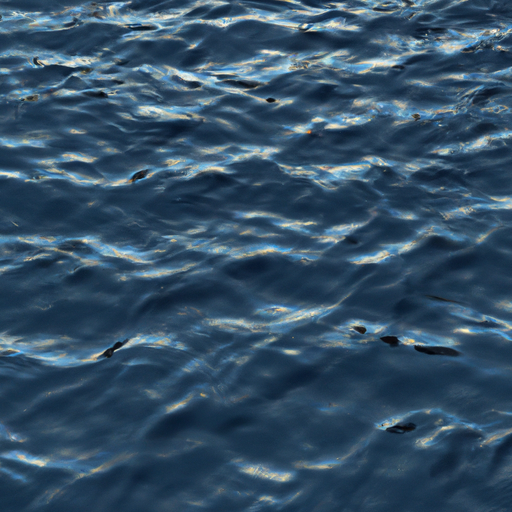} \\
    \includegraphics[width=\linewidth]{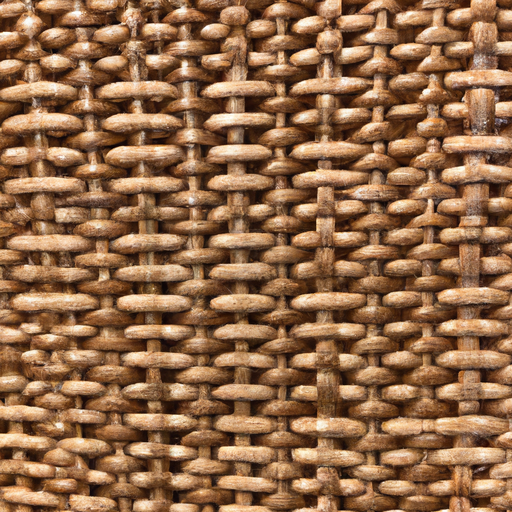} %
    \end{subfigure}
    \begin{subfigure}[t]{0.38\linewidth}
    \includegraphics[width=\linewidth]{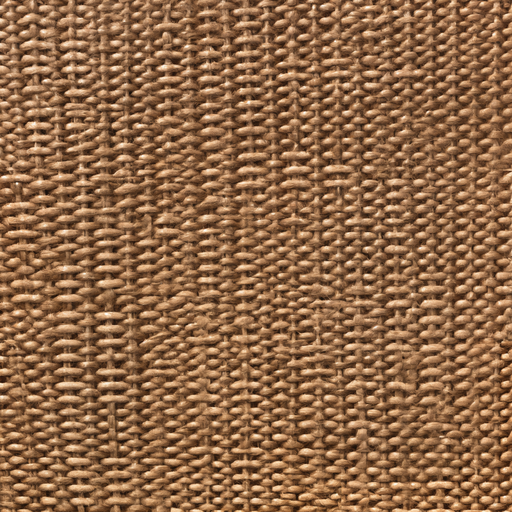} %
    \end{subfigure}

    \caption{\textbf{Limitations}. \model may automatically adjust the lighting, leading to color drift (left). It also may not perform well on textures with strong directional distributions (right). Reference images are shown in the middle column.}
    \label{fig:fail}

    \end{center}
\end{figure}
\section{Conclusion and Discussions}

In this paper, we have presented \model, a method that can generate an infinite number of arbitrarily large, high-quality textures that operates purely from a text prompt. 
This is accomplished through the fine-tuning of a diffusion model and a smart use of the diffusion model at inference time. 
Our diverse range of results demonstrates that the method is capable of synthesizing high-resolution, high-quality textures guided by text prompts. While training a diffusion model per prompt may appear cumbersome, it remains the fastest high-resolution texture generator to our knowledge. 
The trained diffusion model is stable enough for real-world application, enabling photo-realistic renderings for 3D shape collections. 
We have further showcased an application by utilizing synthesized textures for retexturing any given image. We believe that this work represents significant progress towards the goal of generating high-quality graphics assets from natural language descriptions.

\textbf{Limitations.}
\model inherits the limitations of Stable Diffusion and may automatically adjust the lighting, leading to color drift (Fig.~\ref{fig:fail} left). Furthermore, \model may not perform optimally on textures with strong directional distributions (Fig.~\ref{fig:fail} right). This limitation arises from the fact that \model denoises random patches during testing, which may not adequately preserve the underlying direction statistics.

\textbf{Acknowledgements.}
We thank Beibin Li, Philipp Henzler, Dor Verbin, and Richard Szeliski for helpful discussions and feedback.
This work was supported by the UW Reality Lab, Meta, Google, OPPO, and Amazon.

\clearpage

{
    \small
    \bibliographystyle{ieeenat_fullname}
    \bibliography{main}

\begin{thebibliography}{53}
\providecommand{\natexlab}[1]{#1}
\providecommand{\url}[1]{\texttt{#1}}
\expandafter\ifx\csname urlstyle\endcsname\relax
  \providecommand{\doi}[1]{doi: #1}\else
  \providecommand{\doi}{doi: \begingroup \urlstyle{rm}\Url}\fi

\bibitem[Bar{-}Tal et~al.(2023)Bar{-}Tal, Yariv, Lipman, and Dekel]{bar2023multidiffusion}
Omer Bar{-}Tal, Lior Yariv, Yaron Lipman, and Tali Dekel.
\newblock Multidiffusion: Fusing diffusion paths for controlled image generation.
\newblock In \emph{International Conference on Machine Learning, {ICML} 2023, 23-29 July 2023, Honolulu, Hawaii, {USA}}, pages 1737--1752. {PMLR}, 2023.

\bibitem[Bergmann et~al.(2017)Bergmann, Jetchev, and Vollgraf]{bergmann2017learning}
Urs Bergmann, Nikolay Jetchev, and Roland Vollgraf.
\newblock Learning texture manifolds with the periodic spatial gan.
\newblock \emph{arXiv preprint arXiv:1705.06566}, 2017.

\bibitem[Darabi et~al.(2012)Darabi, Shechtman, Barnes, Goldman, and Sen]{darabi2012image}
Soheil Darabi, Eli Shechtman, Connelly Barnes, Dan~B Goldman, and Pradeep Sen.
\newblock Image melding: Combining inconsistent images using patch-based synthesis.
\newblock \emph{ACM Transactions on graphics (TOG)}, 31\penalty0 (4):\penalty0 1--10, 2012.

\bibitem[Dhariwal and Nichol(2021)]{dhariwal2021diffusion}
Prafulla Dhariwal and Alexander Nichol.
\newblock Diffusion models beat gans on image synthesis.
\newblock \emph{Advances in Neural Information Processing Systems}, 34:\penalty0 8780--8794, 2021.

\bibitem[Efros and Freeman(2001)]{efros2001image}
Alexei~A Efros and William~T Freeman.
\newblock Image quilting for texture synthesis and transfer.
\newblock In \emph{Proceedings of the 28th annual conference on Computer graphics and interactive techniques}, pages 341--346, 2001.

\bibitem[Efros and Leung(1999)]{efros1999texture}
Alexei~A Efros and Thomas~K Leung.
\newblock Texture synthesis by non-parametric sampling.
\newblock In \emph{Proceedings of the seventh IEEE international conference on computer vision}, pages 1033--1038. IEEE, 1999.

\bibitem[Gatys et~al.(2015{\natexlab{a}})Gatys, Ecker, and Bethge]{gatys2015texture}
Leon Gatys, Alexander~S Ecker, and Matthias Bethge.
\newblock Texture synthesis using convolutional neural networks.
\newblock \emph{Advances in neural information processing systems}, 28, 2015{\natexlab{a}}.

\bibitem[Gatys et~al.(2015{\natexlab{b}})Gatys, Ecker, and Bethge]{gatys2015neural}
Leon~A Gatys, Alexander~S Ecker, and Matthias Bethge.
\newblock A neural algorithm of artistic style.
\newblock \emph{arXiv preprint arXiv:1508.06576}, 2015{\natexlab{b}}.

\bibitem[Hertzmann et~al.(2001)Hertzmann, Jacobs, Oliver, Curless, and Salesin]{hertzmann2001image}
Aaron Hertzmann, Charles~E Jacobs, Nuria Oliver, Brian Curless, and David~H Salesin.
\newblock Image analogies.
\newblock In \emph{Proceedings of the 28th annual conference on Computer graphics and interactive techniques}, pages 327--340, 2001.

\bibitem[Ho et~al.(2020)Ho, Jain, and Abbeel]{ho2020denoising}
Jonathan Ho, Ajay Jain, and Pieter Abbeel.
\newblock Denoising diffusion probabilistic models.
\newblock \emph{Advances in Neural Information Processing Systems}, 33:\penalty0 6840--6851, 2020.

\bibitem[Jetchev et~al.(2016)Jetchev, Bergmann, and Vollgraf]{jetchev2016texture}
Nikolay Jetchev, Urs Bergmann, and Roland Vollgraf.
\newblock Texture synthesis with spatial generative adversarial networks.
\newblock \emph{arXiv preprint arXiv:1611.08207}, 2016.

\bibitem[Johnson et~al.(2016)Johnson, Alahi, and Fei-Fei]{johnson2016perceptual}
Justin Johnson, Alexandre Alahi, and Li Fei-Fei.
\newblock Perceptual losses for real-time style transfer and super-resolution.
\newblock In \emph{Computer Vision--ECCV 2016: 14th European Conference, Amsterdam, The Netherlands, October 11-14, 2016, Proceedings, Part II 14}, pages 694--711. Springer, 2016.

\bibitem[Kaspar et~al.(2015)Kaspar, Neubert, Lischinski, Pauly, and Kopf]{kaspar2015self}
Alexandre Kaspar, Boris Neubert, Dani Lischinski, Mark Pauly, and Johannes Kopf.
\newblock Self tuning texture optimization.
\newblock In \emph{Computer Graphics Forum}, pages 349--359. Wiley Online Library, 2015.

\bibitem[Kwatra et~al.(2003)Kwatra, Sch{\"o}dl, Essa, Turk, and Bobick]{kwatra2003graphcut}
Vivek Kwatra, Arno Sch{\"o}dl, Irfan Essa, Greg Turk, and Aaron Bobick.
\newblock Graphcut textures: Image and video synthesis using graph cuts.
\newblock \emph{Acm transactions on graphics (tog)}, 22\penalty0 (3):\penalty0 277--286, 2003.

\bibitem[Li et~al.(2017)Li, Fang, Yang, Wang, Lu, and Yang]{li2017diversified}
Yijun Li, Chen Fang, Jimei Yang, Zhaowen Wang, Xin Lu, and Ming-Hsuan Yang.
\newblock Diversified texture synthesis with feed-forward networks.
\newblock In \emph{Proceedings of the IEEE conference on computer vision and pattern recognition}, pages 3920--3928, 2017.

\bibitem[Liang et~al.(2001)Liang, Liu, Xu, Guo, and Shum]{liang2001real}
Lin Liang, Ce Liu, Ying-Qing Xu, Baining Guo, and Heung-Yeung Shum.
\newblock Real-time texture synthesis by patch-based sampling.
\newblock \emph{ACM Transactions on Graphics (ToG)}, 20\penalty0 (3):\penalty0 127--150, 2001.

\bibitem[Lin et~al.(2014)Lin, Maire, Belongie, Hays, Perona, Ramanan, Doll{\'a}r, and Zitnick]{lin2014microsoft}
Tsung-Yi Lin, Michael Maire, Serge Belongie, James Hays, Pietro Perona, Deva Ramanan, Piotr Doll{\'a}r, and C~Lawrence Zitnick.
\newblock Microsoft coco: Common objects in context.
\newblock In \emph{Computer Vision--ECCV 2014: 13th European Conference, Zurich, Switzerland, September 6-12, 2014, Proceedings, Part V 13}, pages 740--755. Springer, 2014.

\bibitem[Lin et~al.(2006)Lin, Hays, Wu, Liu, and Kwatra]{lin2006quantitative}
Wen-Chieh Lin, James Hays, Chenyu Wu, Yanxi Liu, and Vivek Kwatra.
\newblock Quantitative evaluation of near regular texture synthesis algorithms.
\newblock In \emph{2006 IEEE Computer Society Conference on Computer Vision and Pattern Recognition (CVPR'06)}, pages 427--434. IEEE, 2006.

\bibitem[Lu et~al.(2022)Lu, Zhou, Bao, Chen, Li, and Zhu]{lu2022dpm}
Cheng Lu, Yuhao Zhou, Fan Bao, Jianfei Chen, Chongxuan Li, and Jun Zhu.
\newblock Dpm-solver: A fast ode solver for diffusion probabilistic model sampling in around 10 steps.
\newblock \emph{arXiv preprint arXiv:2206.00927}, 2022.

\bibitem[Mansimov et~al.(2015)Mansimov, Parisotto, Ba, and Salakhutdinov]{mansimov2015generating}
Elman Mansimov, Emilio Parisotto, Jimmy~Lei Ba, and Ruslan Salakhutdinov.
\newblock Generating images from captions with attention.
\newblock \emph{arXiv preprint arXiv:1511.02793}, 2015.

\bibitem[Meng et~al.(2022)Meng, Gao, Kingma, Ermon, Ho, and Salimans]{meng2022distillation}
Chenlin Meng, Ruiqi Gao, Diederik~P Kingma, Stefano Ermon, Jonathan Ho, and Tim Salimans.
\newblock On distillation of guided diffusion models.
\newblock \emph{arXiv preprint arXiv:2210.03142}, 2022.

\bibitem[Mordvintsev and Niklasson(2021)]{mordvintsev2021mu}
Alexander Mordvintsev and Eyvind Niklasson.
\newblock $\mu$nca: Texture generation with ultra-compact neural cellular automata.
\newblock \emph{arXiv preprint arXiv:2111.13545}, 2021.

\bibitem[Nichol et~al.(2021)Nichol, Dhariwal, Ramesh, Shyam, Mishkin, McGrew, Sutskever, and Chen]{nichol2021glide}
Alex Nichol, Prafulla Dhariwal, Aditya Ramesh, Pranav Shyam, Pamela Mishkin, Bob McGrew, Ilya Sutskever, and Mark Chen.
\newblock Glide: Towards photorealistic image generation and editing with text-guided diffusion models.
\newblock \emph{arXiv preprint arXiv:2112.10741}, 2021.

\bibitem[Niklasson et~al.(2021)Niklasson, Mordvintsev, Randazzo, and Levin]{niklasson2021self}
Eyvind Niklasson, Alexander Mordvintsev, Ettore Randazzo, and Michael Levin.
\newblock Self-organising textures.
\newblock \emph{Distill}, 6\penalty0 (2):\penalty0 e00027--003, 2021.

\bibitem[Praun et~al.(2000)Praun, Finkelstein, and Hoppe]{praun2000lapped}
Emil Praun, Adam Finkelstein, and Hugues Hoppe.
\newblock Lapped textures.
\newblock In \emph{Proceedings of the 27th annual conference on Computer graphics and interactive techniques}, pages 465--470, 2000.

\bibitem[Radford et~al.(2015)Radford, Metz, and Chintala]{radford2015unsupervised}
Alec Radford, Luke Metz, and Soumith Chintala.
\newblock Unsupervised representation learning with deep convolutional generative adversarial networks.
\newblock \emph{arXiv preprint arXiv:1511.06434}, 2015.

\bibitem[Radford et~al.(2021)Radford, Kim, Hallacy, Ramesh, Goh, Agarwal, Sastry, Askell, Mishkin, Clark, et~al.]{radford2021learning}
Alec Radford, Jong~Wook Kim, Chris Hallacy, Aditya Ramesh, Gabriel Goh, Sandhini Agarwal, Girish Sastry, Amanda Askell, Pamela Mishkin, Jack Clark, et~al.
\newblock Learning transferable visual models from natural language supervision.
\newblock In \emph{International conference on machine learning}, pages 8748--8763. PMLR, 2021.

\bibitem[Raffel et~al.(2020)Raffel, Shazeer, Roberts, Lee, Narang, Matena, Zhou, Li, and Liu]{raffel2020exploring}
Colin Raffel, Noam Shazeer, Adam Roberts, Katherine Lee, Sharan Narang, Michael Matena, Yanqi Zhou, Wei Li, and Peter~J Liu.
\newblock Exploring the limits of transfer learning with a unified text-to-text transformer.
\newblock \emph{The Journal of Machine Learning Research}, 21\penalty0 (1):\penalty0 5485--5551, 2020.

\bibitem[Ramesh et~al.(2022)Ramesh, Dhariwal, Nichol, Chu, and Chen]{ramesh2022hierarchical}
Aditya Ramesh, Prafulla Dhariwal, Alex Nichol, Casey Chu, and Mark Chen.
\newblock Hierarchical text-conditional image generation with clip latents.
\newblock \emph{arXiv preprint arXiv:2204.06125}, 2022.

\bibitem[Ranftl et~al.(2022)Ranftl, Lasinger, Hafner, Schindler, and Koltun]{Ranftl2022}
Ren\'{e} Ranftl, Katrin Lasinger, David Hafner, Konrad Schindler, and Vladlen Koltun.
\newblock Towards robust monocular depth estimation: Mixing datasets for zero-shot cross-dataset transfer.
\newblock \emph{IEEE Transactions on Pattern Analysis and Machine Intelligence}, 44\penalty0 (3), 2022.

\bibitem[Reed et~al.(2016)Reed, Akata, Yan, Logeswaran, Schiele, and Lee]{reed2016generative}
Scott Reed, Zeynep Akata, Xinchen Yan, Lajanugen Logeswaran, Bernt Schiele, and Honglak Lee.
\newblock Generative adversarial text to image synthesis.
\newblock In \emph{International conference on machine learning}, pages 1060--1069. PMLR, 2016.

\bibitem[Rombach et~al.(2021)Rombach, Blattmann, Lorenz, Esser, and Ommer]{rombach2021highresolution}
Robin Rombach, Andreas Blattmann, Dominik Lorenz, Patrick Esser, and Björn Ommer.
\newblock High-resolution image synthesis with latent diffusion models, 2021.

\bibitem[Ruiz et~al.(2023)Ruiz, Li, Jampani, Pritch, Rubinstein, and Aberman]{ruiz2022dreambooth}
Nataniel Ruiz, Yuanzhen Li, Varun Jampani, Yael Pritch, Michael Rubinstein, and Kfir Aberman.
\newblock Dreambooth: Fine tuning text-to-image diffusion models for subject-driven generation.
\newblock In \emph{Proceedings of the IEEE/CVF Conference on Computer Vision and Pattern Recognition}, pages 22500--22510, 2023.

\bibitem[Saharia et~al.(2022)Saharia, Chan, Saxena, Li, Whang, Denton, Ghasemipour, Gontijo~Lopes, Karagol~Ayan, Salimans, et~al.]{saharia2022photorealistic}
Chitwan Saharia, William Chan, Saurabh Saxena, Lala Li, Jay Whang, Emily~L Denton, Kamyar Ghasemipour, Raphael Gontijo~Lopes, Burcu Karagol~Ayan, Tim Salimans, et~al.
\newblock Photorealistic text-to-image diffusion models with deep language understanding.
\newblock \emph{Advances in Neural Information Processing Systems}, 35:\penalty0 36479--36494, 2022.

\bibitem[Salimans and Ho(2022)]{salimans2022progressive}
Tim Salimans and Jonathan Ho.
\newblock Progressive distillation for fast sampling of diffusion models.
\newblock \emph{arXiv preprint arXiv:2202.00512}, 2022.

\bibitem[Song et~al.(2020)Song, Meng, and Ermon]{song2020denoising}
Jiaming Song, Chenlin Meng, and Stefano Ermon.
\newblock Denoising diffusion implicit models.
\newblock \emph{arXiv preprint arXiv:2010.02502}, 2020.

\bibitem[Tao et~al.(2022)Tao, Tang, Wu, Jing, Bao, and Xu]{tao2022df}
Ming Tao, Hao Tang, Fei Wu, Xiao-Yuan Jing, Bing-Kun Bao, and Changsheng Xu.
\newblock Df-gan: A simple and effective baseline for text-to-image synthesis.
\newblock In \emph{Proceedings of the IEEE/CVF Conference on Computer Vision and Pattern Recognition}, pages 16515--16525, 2022.

\bibitem[Turk(1991)]{turk1991generating}
Greg Turk.
\newblock Generating textures on arbitrary surfaces using reaction-diffusion.
\newblock \emph{Acm Siggraph Computer Graphics}, 25\penalty0 (4):\penalty0 289--298, 1991.

\bibitem[Turk(2001)]{turk2001texture}
Greg Turk.
\newblock Texture synthesis on surfaces.
\newblock In \emph{Proceedings of the 28th annual conference on Computer graphics and interactive techniques}, pages 347--354, 2001.

\bibitem[Ulyanov et~al.(2016)Ulyanov, Lebedev, Vedaldi, and Lempitsky]{ulyanov2016texture}
Dmitry Ulyanov, Vadim Lebedev, Andrea Vedaldi, and Victor Lempitsky.
\newblock Texture networks: Feed-forward synthesis of textures and stylized images.
\newblock \emph{arXiv preprint arXiv:1603.03417}, 2016.

\bibitem[Ulyanov et~al.(2017)Ulyanov, Vedaldi, and Lempitsky]{ulyanov2017improved}
Dmitry Ulyanov, Andrea Vedaldi, and Victor Lempitsky.
\newblock Improved texture networks: Maximizing quality and diversity in feed-forward stylization and texture synthesis.
\newblock In \emph{Proceedings of the IEEE conference on computer vision and pattern recognition}, pages 6924--6932, 2017.

\bibitem[Verbin and Zickler(2020)]{Verbin_2020_CVPR}
Dor Verbin and Todd Zickler.
\newblock Toward a universal model for shape from texture.
\newblock In \emph{Proceedings of the IEEE/CVF Conference on Computer Vision and Pattern Recognition (CVPR)}, 2020.

\bibitem[Verbin et~al.(2021)Verbin, Gortler, and Zickler]{verbin2}
Dor Verbin, Steven~J. Gortler, and Todd Zickler.
\newblock Unique geometry and texture from corresponding image patches.
\newblock \emph{IEEE Transactions on Pattern Analysis and Machine Intelligence}, 43\penalty0 (12):\penalty0 4519--4522, 2021.

\bibitem[Wang et~al.(2022)Wang, Zhang, Zhang, Ouyang, Chen, Chen, and Wen]{wang2022pretraining}
Tengfei Wang, Ting Zhang, Bo Zhang, Hao Ouyang, Dong Chen, Qifeng Chen, and Fang Wen.
\newblock Pretraining is all you need for image-to-image translation.
\newblock \emph{arXiv preprint arXiv:2205.12952}, 2022.

\bibitem[Wei and Levoy(2000)]{wei2000fast}
Li-Yi Wei and Marc Levoy.
\newblock Fast texture synthesis using tree-structured vector quantization.
\newblock In \emph{Proceedings of the 27th annual conference on Computer graphics and interactive techniques}, pages 479--488, 2000.

\bibitem[Welinder et~al.(2011)Welinder, Branson, Mita, Wah, Schroff, Belongie, and Perona]{welinder2010caltech}
Peter Welinder, Steve Branson, Takeshi Mita, Catherine Wah, Florian Schroff, Serge Belongie, and Pietro Perona.
\newblock \emph{Caltech-UCSD Birds 200}.
\newblock 2011.

\bibitem[Witkin and Kass(1991)]{witkin1991reaction}
Andrew Witkin and Michael Kass.
\newblock Reaction-diffusion textures.
\newblock In \emph{Proceedings of the 18th annual conference on computer graphics and interactive techniques}, pages 299--308, 1991.

\bibitem[Xu et~al.(2018)Xu, Zhang, Huang, Zhang, Gan, Huang, and He]{xu2018attngan}
Tao Xu, Pengchuan Zhang, Qiuyuan Huang, Han Zhang, Zhe Gan, Xiaolei Huang, and Xiaodong He.
\newblock Attngan: Fine-grained text to image generation with attentional generative adversarial networks.
\newblock In \emph{Proceedings of the IEEE conference on computer vision and pattern recognition}, pages 1316--1324, 2018.

\bibitem[Zhang et~al.(2017)Zhang, Xu, Li, Zhang, Wang, Huang, and Metaxas]{zhang2017stackgan}
Han Zhang, Tao Xu, Hongsheng Li, Shaoting Zhang, Xiaogang Wang, Xiaolei Huang, and Dimitris~N Metaxas.
\newblock Stackgan: Text to photo-realistic image synthesis with stacked generative adversarial networks.
\newblock In \emph{Proceedings of the IEEE international conference on computer vision}, pages 5907--5915, 2017.

\bibitem[Zhang et~al.(2023)Zhang, Rao, and Agrawala]{zhang2023adding}
Lvmin Zhang, Anyi Rao, and Maneesh Agrawala.
\newblock Adding conditional control to text-to-image diffusion models.
\newblock In \emph{Proceedings of the IEEE/CVF International Conference on Computer Vision}, pages 3836--3847, 2023.

\bibitem[Zhou et~al.(2018)Zhou, Zhu, Bai, Lischinski, Cohen-Or, and Huang]{zhou2018non}
Yang Zhou, Zhen Zhu, Xiang Bai, Dani Lischinski, Daniel Cohen-Or, and Hui Huang.
\newblock Non-stationary texture synthesis by adversarial expansion.
\newblock \emph{arXiv preprint arXiv:1805.04487}, 2018.

\bibitem[Zhu et~al.(2017)Zhu, Park, Isola, and Efros]{zhu2017unpaired}
Jun-Yan Zhu, Taesung Park, Phillip Isola, and Alexei~A Efros.
\newblock Unpaired image-to-image translation using cycle-consistent adversarial networks.
\newblock In \emph{Proceedings of the IEEE international conference on computer vision}, pages 2223--2232, 2017.

\bibitem[Zhu et~al.(2019)Zhu, Pan, Chen, and Yang]{zhu2019dm}
Minfeng Zhu, Pingbo Pan, Wei Chen, and Yi Yang.
\newblock Dm-gan: Dynamic memory generative adversarial networks for text-to-image synthesis.
\newblock In \emph{Proceedings of the IEEE/CVF conference on computer vision and pattern recognition}, pages 5802--5810, 2019.

\end{thebibliography}
}

\end{document}